\documentclass{article} 
\usepackage{colm2024_conference}

\usepackage{microtype}
\usepackage{hyperref}
\usepackage{url}
\usepackage{booktabs}
\usepackage{graphicx}
\usepackage{subfigure}
\usepackage{amsmath}
\usepackage{siunitx}
\usepackage{booktabs}
\usepackage{xcolor}
\usepackage{makecell}
\usepackage{listings}
\usepackage{enumitem}
\usepackage{array}
\usepackage{wrapfig}
\usepackage{multirow}
\usepackage{colortbl}

\newif\ifcomments
\commentstrue 
\ifcomments
    \providecommand{\kai}[1]{{\protect\color{blue}{[kai: #1]}}}
    \providecommand{\eric}[1]{{\protect\color{magenta}{[Eric: #1]}}}
    \providecommand{\reimar}[1]{{\protect\color{purple}{\bf [reimar: #1]}}}
    \providecommand{\alex}[1]{{\protect\color{purple!50!orange}{[alex: #1]}}}
    \providecommand{\lilian}[1]{{\protect\color{purple}{\bf [lilian: #1]}}}
    \providecommand{\johannes}[1]{{\protect\color{purple}{\bf [johannes: #1]}}}
\else
    \providecommand{\kai}[1]{}
    \providecommand{\eric}[1]{}
    \providecommand{\reimar}[1]{}
    \providecommand{\alex}[1]{}
    \providecommand{\lilian}[1]{}
    \providecommand{\johannes}[1]{}
\fi

{\raggedright
\title{Predicting Emergent Capabilities by Finetuning}}

\author{Charlie Snell ~~~~~~ Eric Wallace ~~~~~~ Dan Klein ~~~~~~ Sergey Levine \\
University of California, Berkeley
}

\colmfinalcopy 
\begin{document}

\maketitle

\begin{abstract}
A fundamental open challenge in modern LLM scaling is the lack of understanding around emergent capabilities.
In particular, language model pretraining loss is known to be highly predictable as a function of compute. However, downstream capabilities are far less predictable---sometimes even exhibiting emergent jumps---which makes it challenging to anticipate the capabilities of future models.
In this work, we first pose the task of emergence prediction: given access to current LLMs that have random few-shot accuracy on a task, can we predict whether future models (GPT-N+1) will have non-trivial accuracy on that task?
We then discover a simple insight for this problem: \textit{finetuning LLMs on a given task can shift the point in scaling at which emergence occurs towards less capable models}.
To operationalize this insight, we can finetune LLMs with varying amounts of data and fit a parametric function that predicts when emergence will occur (i.e., ``emergence laws'').
We validate this approach using four standard NLP benchmarks where large-scale open-source LLMs already demonstrate emergence (MMLU, GSM8K, CommonsenseQA, and CoLA). Using only small-scale LLMs, we find that, in some cases, we can accurately predict whether models trained with up to 4x more compute have emerged. Finally, we present a case study of two realistic uses for emergence prediction.

\end{abstract}

\let\originalthefootnote\thefootnote
\let\thefootnote\relax\footnotetext{\scriptsize{Corresponding author: csnell22@berkeley.edu. An early version of this work appeared in COLM 2024.}}
\let\thefootnote\originalthefootnote

\section{Introduction}

The pretraining loss for language models has been shown to follow a simple predictable power law as a function of compute, model parameters, and data~\citep{kaplan2020scaling,hoffmann2022training}. This finding has enabled a precise empirical science to develop around the scaling behavior of LLMs~\citep{muennighoff2023scaling,aghajanyan2023scaling,krajewski2024scaling}, which has in turn led to much of the rapid improvement of language model capabilities in recent years~\citep{openai2024gpt4,geminiteam2023gemini}. On the other hand, the specific downstream capabilities corresponding to a given pretraining loss are generally much less predictable, posing significant challenges for 1) model developers who may want to make specific architectural or data decisions on the basis of future LLM capabilities; 2) policymakers who will need time to assess, plan for, and prepare for potentially dangerous future LLM capabilities like deception, bio-risk, or malicious software agents~\citep{shevlane2023model,bengio2023managing}; and 3) stakeholders who need to make reliable business, financial, and investment decisions on the basis of future LLM capabilities.

\begin{figure*}
    \centering
    \includegraphics[width=0.99\linewidth]{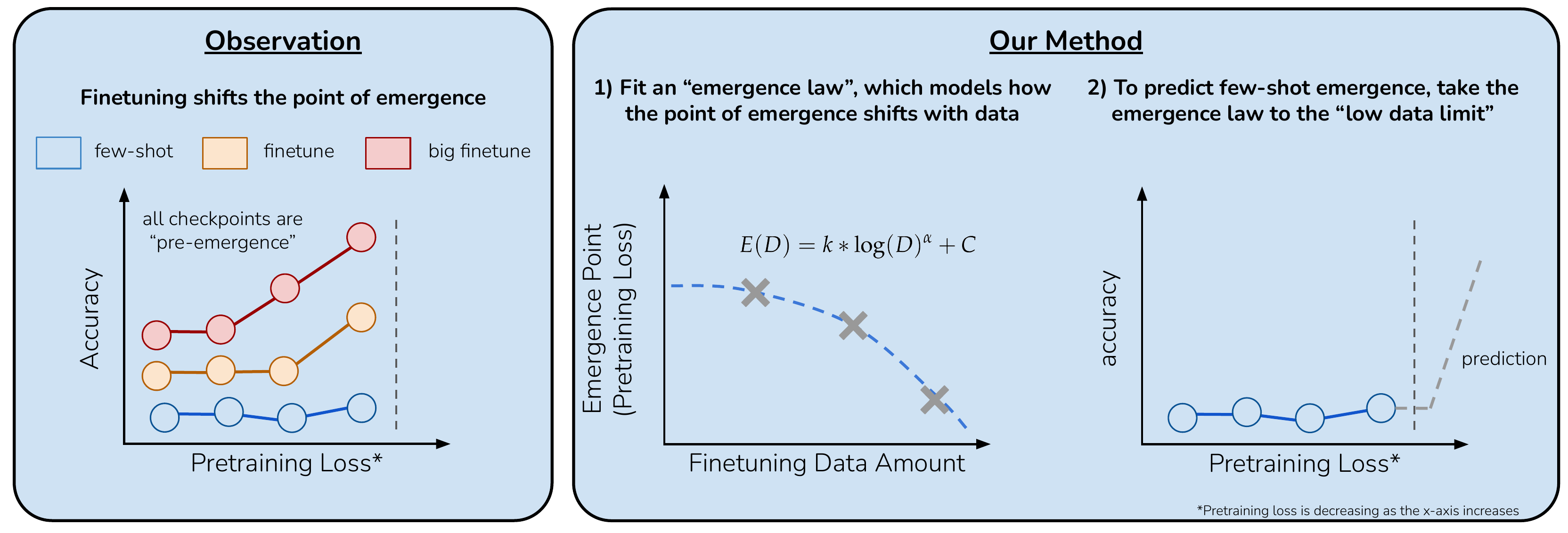}
    \caption{\footnotesize{We find that \textbf{task-specific finetuning systematically shifts the point of emergence towards less capable models}. Motivated by this finding, we develop an \textbf{\emph{emergence law}}, which models how the point of emergence shifts as a function of the amount of finetuning data. Using this emergence law, we can then extrapolate a prediction for the point of emergence in the few-shot setting.}}
    \label{fig:graphical_fig1}
\end{figure*}

Of particular concern and interest is the phenomenon of emergent capabilities in large language models~\cite{wei2022emergent}.
On certain downstream tasks, models may exhibit ``emergence'' wherein only beyond a certain, seemingly arbitrary, threshold in LLM scaling (i.e., the point of emergence) do models spontaneously improve beyond random-chance. In cases where existing models have already crossed the point of emergence on a given task, demonstrating smooth performance improvements (\emph{the post-emergence regime}), it is possible to make highly accurate predictions about the performance of future models~\citep{caballero2023broken,gadre2024language,openai2024gpt4,owen2024predictablelanguagemodelbenchmark}. However, on tasks in which all existing models demonstrate random-chance performance (\emph{the pre-emergence regime}), making any kind of prediction at all about future model capabilities remains an important unsolved challenge in LLM scaling. In this case, there is no known method for predicting at what point, if any, models will demonstrate emergence, let alone how performance will scale thereafter. To this end, we pose the problem of emergence prediction. Concretely, can we accurately predict the point in scaling at which emergence will occur on a given task, while only having access to pre-emergence model checkpoints?

\begin{figure*}
    \centering
    \includegraphics[width=0.99\textwidth]{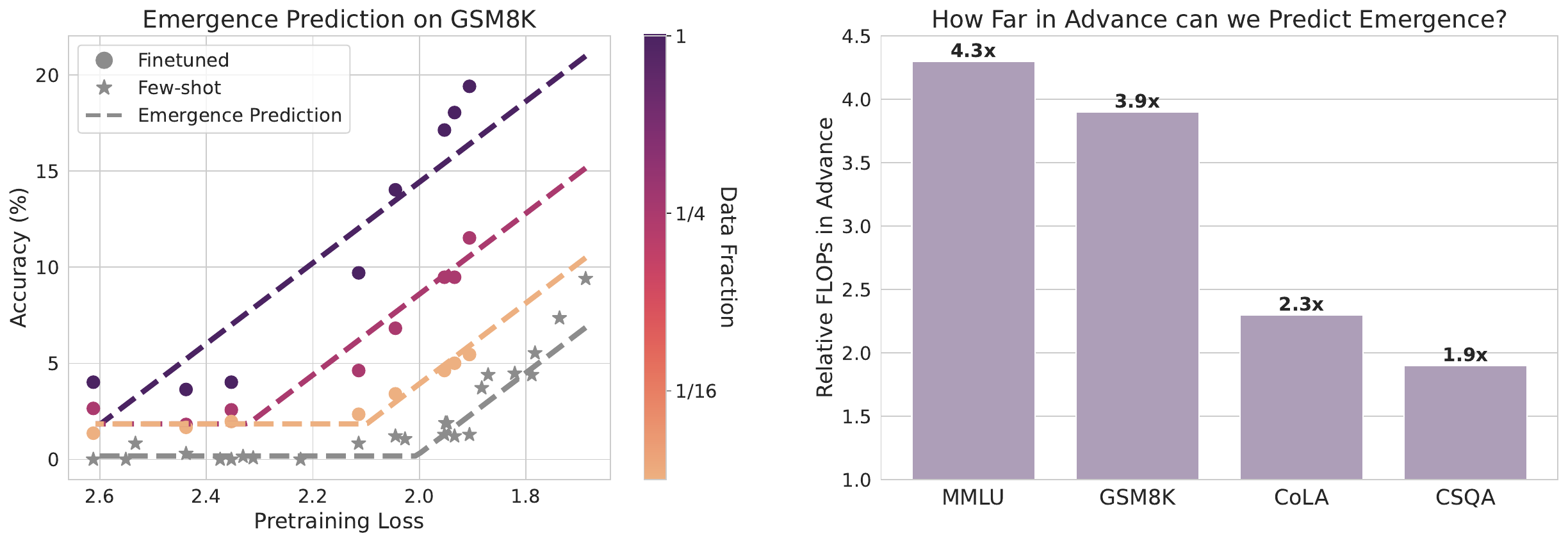}
    \vspace{-0.45cm}
    \caption{\footnotesize{\textbf{Left: we predict emergence in the few-shot setting by leveraging information about how ``pre-emergence'' models behave after finetuning.} Our key finding is that \textit{finetuning effectively shifts the point of emergence from stronger to weaker models}. Moreover, by varying the amount of finetuning data, the emergence point is shifted accordingly. We can use this finding to predict when few-shot emergence will occur by fitting a parametric function to the results (i.e., emergence law) and then taking a limit. \textbf{Right: using this approach, we can predict emergence up to 4x the FLOPs in advance.}}}
    \label{fig:figure1}
\end{figure*}

We demonstrate an extremely simple yet highly effective solution to this problem; namely that \textbf{it is possible to predict few-shot emergent capabilities in future LLMs (e.g., GPT-N+1) by finetuning today's weaker LLMs (e.g., GPT-N)}.
For example, in Figure~\ref{fig:figure1} (left), we see that finetuning models for GSM8K, rather than prompting them, systematically shifts the point of emergence from stronger to weaker LLMs.
Moreover, by varying the amount of finetuning data, the emergence point is shifted accordingly.
Motivated by this finding, we develop an emergence law; a parametric function that models how the point of emergence shifts as a function of the amount of finetuning data.
Using our emergence law, we can extrapolate a prediction for the point of emergence in the few-shot setting (see Figure~\ref{fig:graphical_fig1}).

To validate our approach, we use four standard NLP benchmarks---MMLU, GSM8K, CommonsenseQA, and CoLA---where large-scale open-source LLMs have already demonstrated emergence.
By fitting an emergence law using only small-scale, pre-emergence LLMs, we find that we are not only able to accurately predict the point in scaling at which emergence will occur with more capable LLMs, but also, in some cases, we can do so using models trained with only \textbf{1/4th the FLOPS} needed to achieve emergence (see Figure~\ref{fig:figure1} right).

Finally, we present a case study of two real world uses for emergence prediction. We first demonstrate that emergence prediction can be used to cheaply assess pretraining data quality. Using the difficult APPS coding benchmark, we then demonstrate that our approach can be used to predict more complex capabilities, closer to those of future frontier models.

\section{Background \& Related Work}

\paragraph{Emergence in Large Language Models.} Emergence~\cite{wei2022emergent} refers to the phenomenon in which beyond a certain point in LLM scaling, models suddenly improve beyond chance performance on a given task.
As a result, despite the fact that pretraining loss can be reliably predicted as a function of model scale~\citep{kaplan2020scaling,hoffmann2022training}, the downstream capabilities corresponding to a particular model scale cannot always be.

\paragraph{Practical challenges with emergence.} The phenomenon of emergent capabilities introduces significant challenges for safety preparedness, frontier model development, and business decision making with LLMs. In particular, it is possible that future language models may demonstrate dangerous emergent capabilities, such as planning, deception, bio-risk, or the ability to generate malicious software~\citep{anwar2024foundationalchallengesassuringalignment,hendrycks2023overviewcatastrophicairisks}. Without any ability to predict when and if these capabilities will emerge, we are only left to speculate, which is highly sub-optimal given the potentially high stakes. Furthermore, model developers may need to make architecture and pretraining data decisions on the basis of downstream capabilities which may only emerge with scale, making it challenging and costly to do so. As a result, the inability to reason about and predict emergent capabilities in advance presents a significant unsolved challenge in language model scaling.

\paragraph{Emergence is a Mirage?} A recent work by~\cite{schaeffer2023emergent} claimed that the phenomenon of emergence is not due to sudden fundamental changes in the model, but rather due to the researcher's choice of metric. They argue that emergence is typically observed when using discontinuous metrics of accuracy, such as exact-match, and that using a continuous metric, such as the model's correct answer probability, will instead show smooth scaling. However, in many practical and policy-relevant settings, the metric of primary interest may be fundamentally discontinuous.
Additionally, in some cases finding metrics that yield smooth scaling may be challenging~\citep{barak2023emergent}. In particular, in Figure~\ref{fig:nonsmooth_emergence} we show that when using a continuous probability-based evaluation metric on two canonical language model benchmarks -- namely MMLU~\citep{hendrycks2021measuring} and CommonsenseQA~\citep{talmor2019commonsenseqa} -- we still observe emergence. Similar findings have also been observed by~\cite{du2024understandingemergentabilitieslanguage,lieberum2023doescircuitanalysisinterpretability}. Additional tools are therefore needed to resolve the practical challenges induced by the phenomenon of emergence.

\paragraph{Predicting emergence.} The literature suggests several promising directions for predicting and understanding emergent capabilities.
One direction uses interpretability or training dynamics based techniques to elucidate underlying phase changes inside the model~\citep{lieberum2023doescircuitanalysisinterpretability,olsson2022context}.
Another approach uses a general high-resolution metric which is more amenable to standard scaling analysis~\citep{openai2024gpt4,hu2024predictingemergentabilitiesinfinite}.
Finally, a recent work from~\cite{ruan2024observationalscalinglawspredictability} proposes observational scaling laws, which leverage correlations between many different model families to enable more effective predictions.
Each of these works can be used in complement to our approach: a scaling law inspired ``emergence law'', which predicts emergence by using information about how pre-emergence models behave after being finetuned for a task of interest.

\section{Emergence Prediction}

We now introduce the problem of emergence prediction. We define emergence prediction as the problem of identifying the point in scaling at which emergence will occur using only pre-emergence model checkpoints (i.e., have near-trivial performance on a given task).

\paragraph{Emergence with loss.} We define ``model scale'' to be a model's pretraining loss rather its FLOPs or parameter count. Prior work has found pretraining loss to be highly predictive of downstream capabilities~\citep{du2024understandingemergentabilitieslanguage,gadre2024language,huang2024compressionrepresentsintelligencelinearly,xia2023trainingtrajectorieslanguagemodels}. Thus, it makes for a more precise independent variable when studying emergence.

\begin{figure}
    \centering
    \subfigure{
        \includegraphics[width=0.99\textwidth]{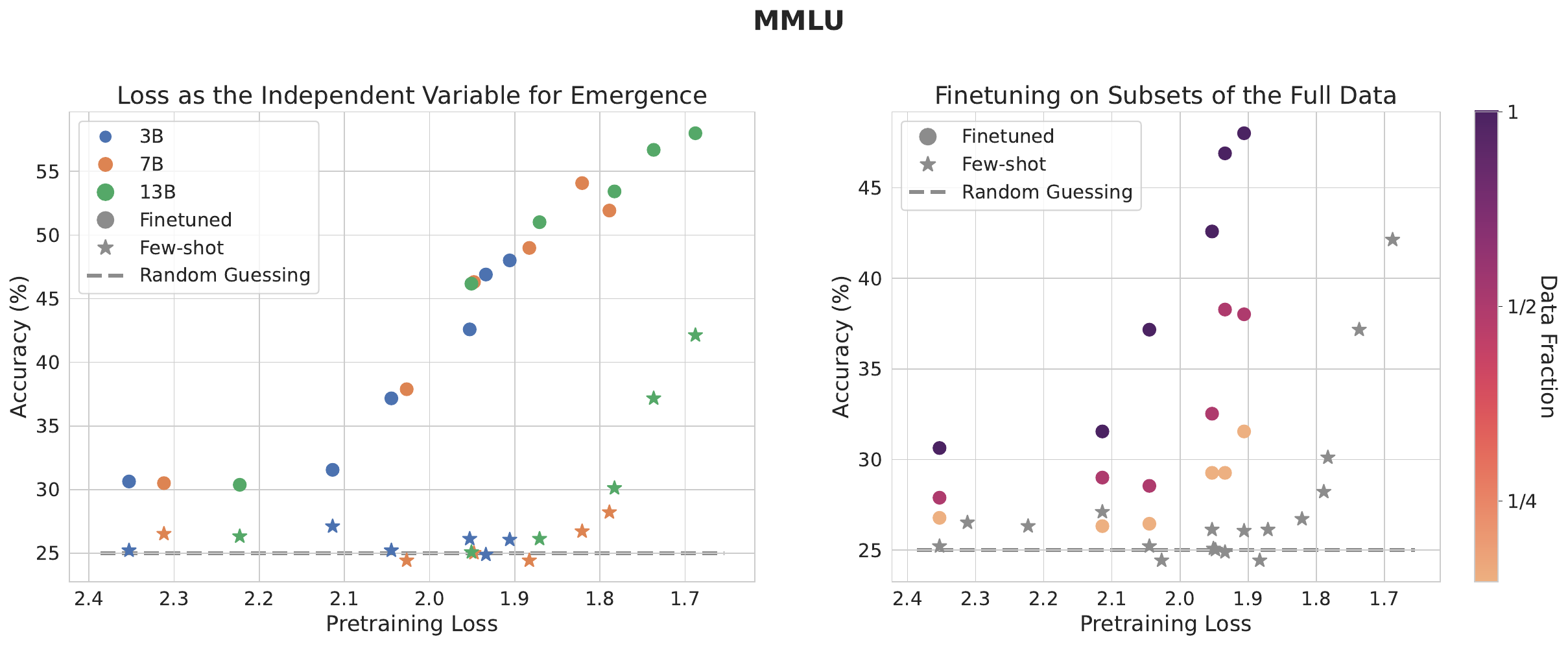}
        \label{fig:mmlu_loss_x_axis_and_finetune_subsets}
    } \\
    \subfigure{
        \includegraphics[width=0.99\textwidth]{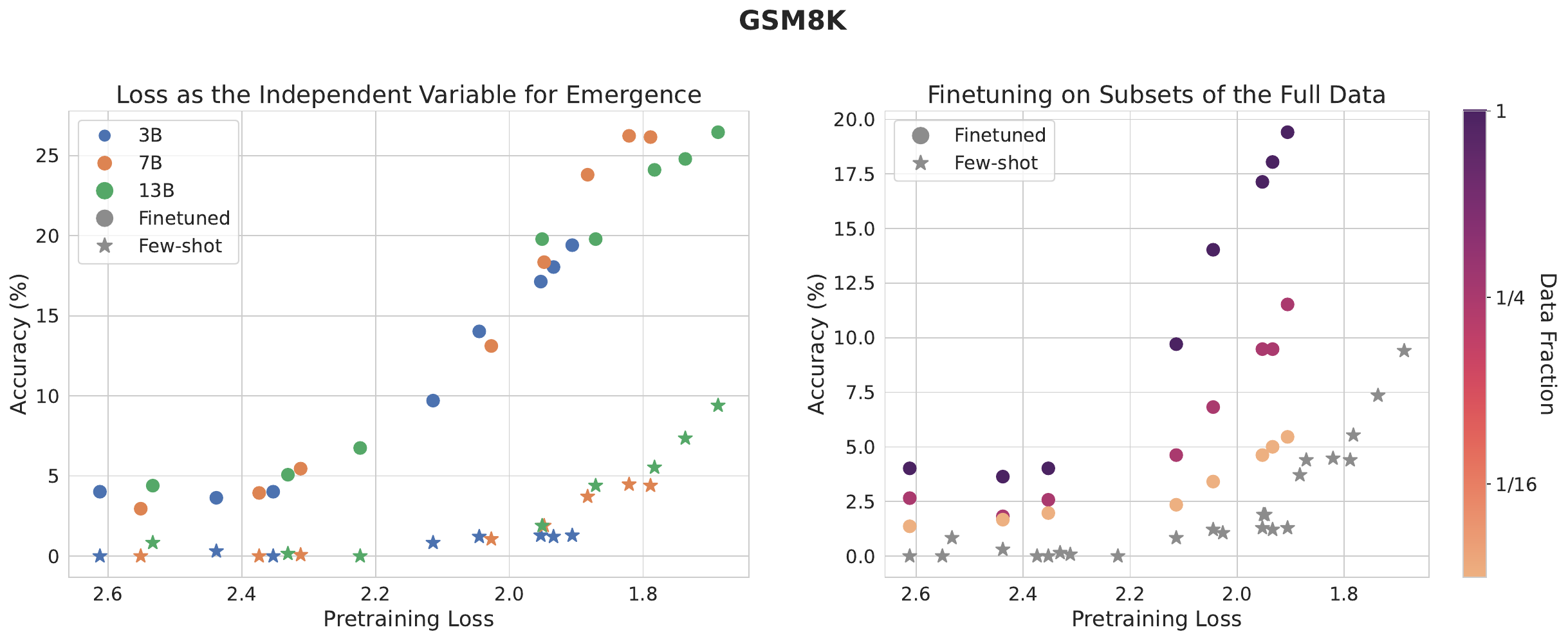}
        \label{fig:gsm8k_loss_x_axis_and_finetune_subsets}
    }
    \caption{\footnotesize{\textbf{Left: the finetuned and few-shot performance of intermediate LLM checkpoints.} We plot downstream accuracy against pretraining loss for all 3B, 7B, and 13B intermediate OpenLLaMA V1 checkpoints on MMLU and GSM8K. We see that the point of emergence is systematically shifted towards weaker LLMs after finetuning. Additionally, the magnitude of the shift is consistent across all model sizes at the same pretraining loss. \textbf{Right: varying the amount of finetuning data.} We finetune the 3B intermediate checkpoints on subsets of the full finetuning data. We see that as we increase the amount of finetuning data, the point of emergence shifts further towards less capable LLMs.}}
    \label{fig:loss_x_axis_and_finetune_subsets}
\end{figure}

\paragraph{Modeling emergence with a ReLU.} To formally define emergence prediction, we model the phenomenon of emergence using a ReLU function. While prior work~\cite{gadre2024language} has found downstream performance to be well modeled by an exponential function of pretraining loss, we use a ReLU because our primary focus is on modeling the point at which emergence occurs, rather than accurately extrapolating performance post-emergence. The ReLU elbow, in this case, provides a clear denotation for the precise point at which emergence begins. Specifically, we model emergence with:

\begin{equation}
    \text{Perf}_{A,B,E}(L(\text{M})) = A * \max(E - L(\text{M}), 0) + B
    \label{eq:relu}
\end{equation}

The output is downstream performance on our task of interest $\text{Perf}(L(\text{M}))$, and the input $L(\text{M})$ corresponds to the pretraining loss of a given model checkpoint ``$\text{M}$''. $A$ and $B$ are parameters defining the slope and floor of the ReLU respectively; and $E$ denotes the point in pretraining loss at which emergence occurs (e.g., the ReLU elbow). The goal of emergence prediction is therefore to accurately predict $E$ for a given task and series of models.

\paragraph{Estimating uncertainty for emergence prediction.} Making an exact pointwise prediction for $E$ may not always be feasible: due to noise in both the model checkpoints and the prediction method, we may have considerable uncertainty about the exact point of emergence. We therefore not only aim to predict a single MLE point for when emergence will occur, but also a calibrated probability distribution over all possible points at which it may occur.

\section{How does Finetuneing Interact with Emergence?}
\label{sec:finetuning_interacts}

In this work, we explore one particular approach to the problem of emergence prediction. Namely, we aim to use information about how pre-emergence models behave under the influence of task-specific finetuning to obtain predictive power about the point of emergence in the few-shot setting. To develop such an approach, we first need to understand how finetuning effects the emergence curve. Here, we empirically characterize this relationship.

\subsection{Experimental Setup}

We conduct experiments on two canonical LLM tasks -- MMLU~\citep{hendrycks2021measuring} and GSM8K~\citep{cobbe2021training} -- in which we observe emergence in the few-shot setting with our most capable open LLMs. We also include results on CommonsenseQA~\citep{talmor2019commonsenseqa} (CSQA) and CoLA~\citep{wang-etal-2018-glue} in Appendix~\ref{app:finetuning_fewshot_across_sizes} and Appendix~\ref{app:full_plots}. To obtain a granular set of measurements along the emergence curve, we not only use different sized models but also different intermediate checkpoints for each model size. Specifically, we use a series of intermediate 3B, 7B, and 13B checkpoints from the OpenLLaMA V1 series~\citep{openlm2023openllama}. For all tasks except MMLU, we finetune on the provided training split, and we evaluate on the test split. For MMLU, there is no formal training set, so we finetune on the test set and evaluate on the validation set. See Appendix~\ref{app:details} for more details.

To understand how the emergence curve is effected by finetuning, we finetune each 3B, 7B, and 13B checkpoint on all tasks. We also finetune all 3B model checkpoints on sampled subsets of the full finetuning data to understand the effect of the amount of data. Finally, to help elucidate the underlying mechanisms behind our observations, we conduct additional experiments with PEFT in Appendix~\ref{app:peft}. In this section, we use full parameter finetuning.

\subsection{Empirical findings}

\paragraph{Finetuning shifts the point of emergence towards weaker LLMs.} In Figure~\ref{fig:loss_x_axis_and_finetune_subsets} (left), we plot the few-shot and finetuned performance of each model against pretraining loss on GSM8K and MMLU. We see that the finetuned models follow a similar emergence ReLU shape as in the few-shot setting. However the finetuned ReLU elbow is systematically shifted towards less capable LLMs. Moreover, \textbf{the shift is consistent across all model sizes} at the same pretraining loss, demonstrating that pretraining loss can act as an effective independent variable for representing capabilities in both the few-shot~\cite{du2024understandingemergentabilitieslanguage,gadre2024language,huang2024compressionrepresentsintelligencelinearly,xia2023trainingtrajectorieslanguagemodels} and finetuned settings, further motivating our use of pretraining loss as the independent variable for emergence prediction.

\paragraph{The emergence shift is effected by the amount of finetuning data.} We also plot the performance of our 3B model checkpoints after being finetuned on subsets of the full data in Figure~\ref{fig:loss_x_axis_and_finetune_subsets} (right). On both MMLU and GSM8K, we see that \textbf{as we increase the amount of finetuning data, the point of point of emergence is shifted further towards less capable LLMs}. The amount of finetuning data can therefore modulate the emergence shift.

These observations demonstrate a clear pattern of structure in how task-specific finetuning interacts with emergence: fine-tuning shifts the emergence elbow from strong to weak models and the amount of finetuning data controls the magnitude of this shift. Next, we will use this insight to develop a scaling-law-style functional form which can effectively predict the point of emergence in the few-shot setting.

\section{Scaling Laws for Emergence Prediction}
\label{sec:scaling_laws_for_pred}

\begin{wraptable}{r}{0.475\textwidth}
\centering
\scriptsize
\setlength{\extrarowheight}{2pt}
\renewcommand{\arraystretch}{1.3}
\begin{tabular}{|>{\centering\arraybackslash}c|>{\raggedright\arraybackslash}p{4.3cm}|}
\hline
\rowcolor{gray!10}
\textbf{Symbol} & \textbf{Description} \\
\hline
$D$ & Amount of finetuning data \\
\hline
$L(\text{M})$ & Pretraining loss of model $\text{M}$ \\
\hline
$\text{Perf}$ & Downstream performance \\
\hline
$A,B,E$ & ReLU parameters \\
\hline
$E_{\theta}(D)$ & Emergence law; models emergence shift \\
\hline
$k, \alpha, C$ & Emergence law parameters \\
\hline
$D_0$ & Low data extrapolation limit \\
\hline
$\Delta$ & Optional parameter; shifts ReLU base \\
\hline
\end{tabular}
\caption{\footnotesize{Symbols used in Section~\ref{sec:scaling_laws_for_pred}.}}
\end{wraptable}

We now use our observations about how finetuning interacts with emergence to develop a novel emergence prediction method.
Specifically, we use the empirical results of many finetuning runs to fit a scaling law inspired parametric function (i.e., an ``emergence law''), which effectively models how the point of emergence shifts as a function of the amount of finetuning data for a task.
By then extrapolating the resulting emergence law into the low data limit, we can make a prediction about the point of emergence in the few-shot setting.
In the following section, we detail our emergence law functional form and how its parameters are fit to empirical data.

\subsection{Modeling the Emergence Shift}
\label{sec:emerg_pred_by_modeling_shift}

\paragraph{Predicting emergence with an emergence law.}
To predict the point of emergence, we use empirical data to fit the parameters of an ``emergence law'': a function $E_{\theta}(D)$ with parameters $\theta$, which models how the emergence ReLU elbow $E$ shifts as a function of the amount of finetuning data $D$. Using our emergence law fit, we can then extrapolate a prediction for the point of emergence in the few-shot setting. In particular, we take a limit to $D_0$ under the model, where $D_0$ is a ``low data extrapolation limit'', representing the few-shot setting. Finally, we can obtain an uncertainty interval for the point of emergence by applying statistical uncertainty estimation techniques to $\theta$.

\paragraph{Emergence law functional form and implementation.}
In practice, we find that the point of emergence is well modeled by a power law in the log of the amount of finetuning data:

$$E_{k,\alpha,C}(D) = k * \log(D)^\alpha + C$$

where $k$, $\alpha$, and $C$ constitute the learnable parameters $\theta$. To obtain our few-shot prediction under this model, we set the ``low data extrapolation limit'' $D_0$ to the number of examples in our few-shot prompt. We then run MCMC posterior sampling, with a uniform prior over the parameters, to obtain a calibrated uncertainty interval for the point of emergence.

\subsection{Fitting the Emergence Law}
\label{sec:fitting_emerg_law}

We cannot directly fit $E_{\theta}(D)$ to a set of ($D$, $E$) input-output observation tuples, since in practice, we do not observe the point of emergence $E$ directly. Instead, we observe $\text{Perf}(L(\text{M}), D)$: downstream performance as a function of a given model checkpoint $\text{M}$ (represented by its pretraining loss $L(\text{M})$, as a sufficient statistic) and the amount of data $D$ used to finetune it. In Figure~\ref{fig:loss_x_axis_and_finetune_subsets} (right) we plot examples of these raw data tuples on MMLU and GSM8K.

In this setting, the point of emergence $E$ corresponding to a particular $D$ is a statistic of the raw data, which can be computed by modeling the data using the ReLU in Equation~\ref{eq:relu}. Therefore, in order to obtain a fit for $E_{\theta}(D)$ using the raw downstream data, we extend Equation~\ref{eq:relu}, by parameterizing the ReLU elbow with $E_{\theta}(D)$. This gives the function:

$$\text{Perf}_{A,B,\theta}(L(\text{M}), D) = A * \max(E_{\theta}(D) - L(\text{M}), 0) + B$$

We fit all parameters -- $A$, $B$, and $\theta$ -- jointly to the raw empirical data using MSE loss. We then extract the resulting $E_{\theta}(D)$ and take a limit to $D_0$, as described in Section~\ref{sec:emerg_pred_by_modeling_shift}.

\subsection{Collecting Empirical Datapoints}
\label{sec:collecting_empirical_data}

To obtain our empirical data, we need:
\textbf{1)} a set of model checkpoints of varying capabilities (e.g., varying $L(\text{M})$); and \textbf{2)} a range of finetuning data amounts (e.g., varying $D$).

\begin{figure}
    \centering
    \includegraphics[width=0.99\textwidth]{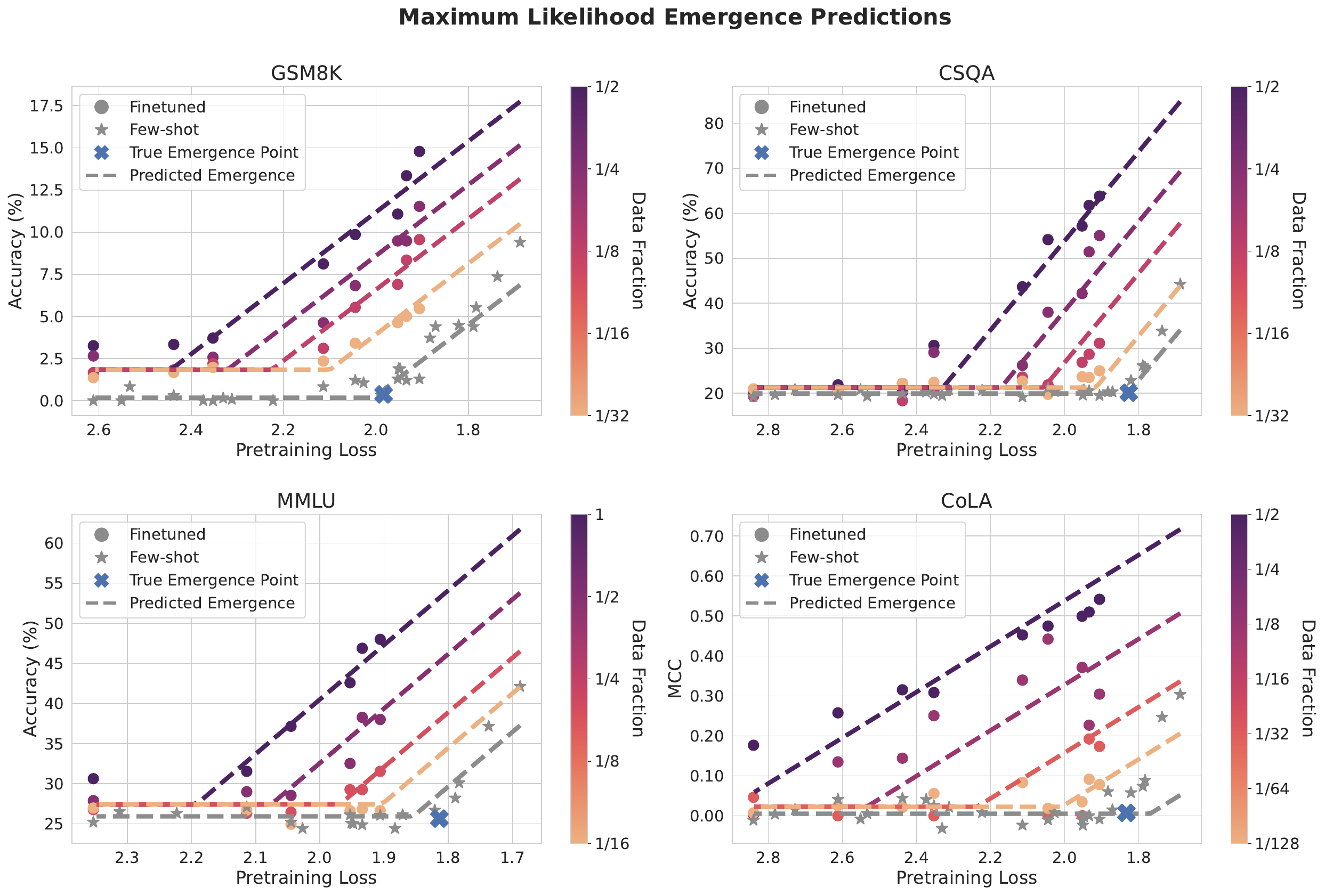}
    \caption{\footnotesize{\textbf{Our MLE emergence law predictions on each task.} The grey line is our extrapolated prediction and the multi-color lines are the fit. While our focus is on predicting the specific point of emergence (e.g., the ReLU elbow), we plot the full ReLU for visual clarity. We see that across all tasks, we are able to successfully predict the point of emergence within 0.1 nats and in many cases much less than that. For visual clarity, we plot a subset of the data used for fitting (see Appendix~\ref{app:full_plots} for all).}}
    \label{fig:fits}
\end{figure}

\paragraph{Selecting the model checkpoints.} To satisfy \textbf{1)}, we can avoid the need to fully pretrain multiple models from scratch by instead utilizing a number of evenly spaced intermediate checkpoints from a single pretraining run. As we observed in Figure~\ref{fig:loss_x_axis_and_finetune_subsets} (left) -- and as noted in recent works~\citep{du2024understandingemergentabilitieslanguage,gadre2024language} -- different-sized models at the same pretraining loss demonstrate similar downstream performance, even after finetuning. Thus, using multiple intermediate checkpoints from the same pretraining run is just as effective as using models from separate pretraining runs of varying scale.

\paragraph{Selecting the finetuning data amounts.}
To obtain a wide range of finetuning data amounts, we randomly sample subsets of the full finetuning data. Since we ultimately want to predict emergence by extrapolating performance into the low data limit, it is important that we focus our data collection on the smallest subsets for which we can still observe emergence with the checkpoints we have. These datapoints will provide some of the most useful signal for accurately extrapolating into the low data limit. In our case, these subsets are often on the order of a few hundred finetuning examples. To further emphasize these datapoints when fitting the emergence law parameters, we weight the MSE loss for each datapoint in an inverse proportion to the amount of finetuning data used.

\subsection{Including Few-shot Data in the Emergence Law}
\label{sec:full_fewshot_form}
In addition to the datapoints obtained from finetuning, we may also optionally want to include the few-shot results from our pre-emergence models when fitting the emergence law. While these few-shot results will all give near-random performance, they will still be informative for telling the model that emergence has not happened yet.

To effectively include these few-shot datapoints in our model, we add an optional parameter $\Delta$, which effectively models the upwards shift in the base of the emergence ReLU that we see when finetuning in Figure~\ref{fig:loss_x_axis_and_finetune_subsets} (right). We believe that this shift is largely due to the model learning trivial features from the fine-tuning data, such as the base-rate of the correct answer\footnote{On GSM8K always guessing the mode answer in the training set gets 2.6\% test accuracy.}. This additional term gives the following model:

$$\text{Perf}_{A, B, \Delta, \theta}(L(\text{M}), D, \mathbb{1}_\text{is finetuned}) = A * \max(E_{\theta}(D) - L(\text{M}), 0) + B + \Delta * \mathbb{1}_\text{is finetuned}$$

Here, $\mathbb{1}_\text{is finetuned}$ indicates if a given datapoint is the result of finetuning, rather than few-shot prompting. For the few-shot datapoints, we set $D$ to $D_0$.

\section{Evaluating the Emergence Law}
\label{sec:evaluating_emergence}

We now evaluate the efficacy of our emergence prediction methodology
on standard NLP benchmarks where large-scale open-source LLMs already demonstrate emergence. In this setting, we fit an emergence law using smaller-scale LLMs, which have random-chance performance on the task. We then check the accuracy of our predictions against the true point of emergence observed from larger models.

In the following section, we first validate that the emergence law presented in Section~\ref{sec:scaling_laws_for_pred} can accurately predict the point of emergence. We then conduct ablations to validate each of our main methodological decisions. Finally, we calculate precisely how far in advance we can predict emergence. We first detail our experimental setup and then discuss each experiment.

\subsection{Experimental Setup}
\label{sec:law_exp_setup}

\begin{figure}
    \centering
    \includegraphics[width=0.99\textwidth]{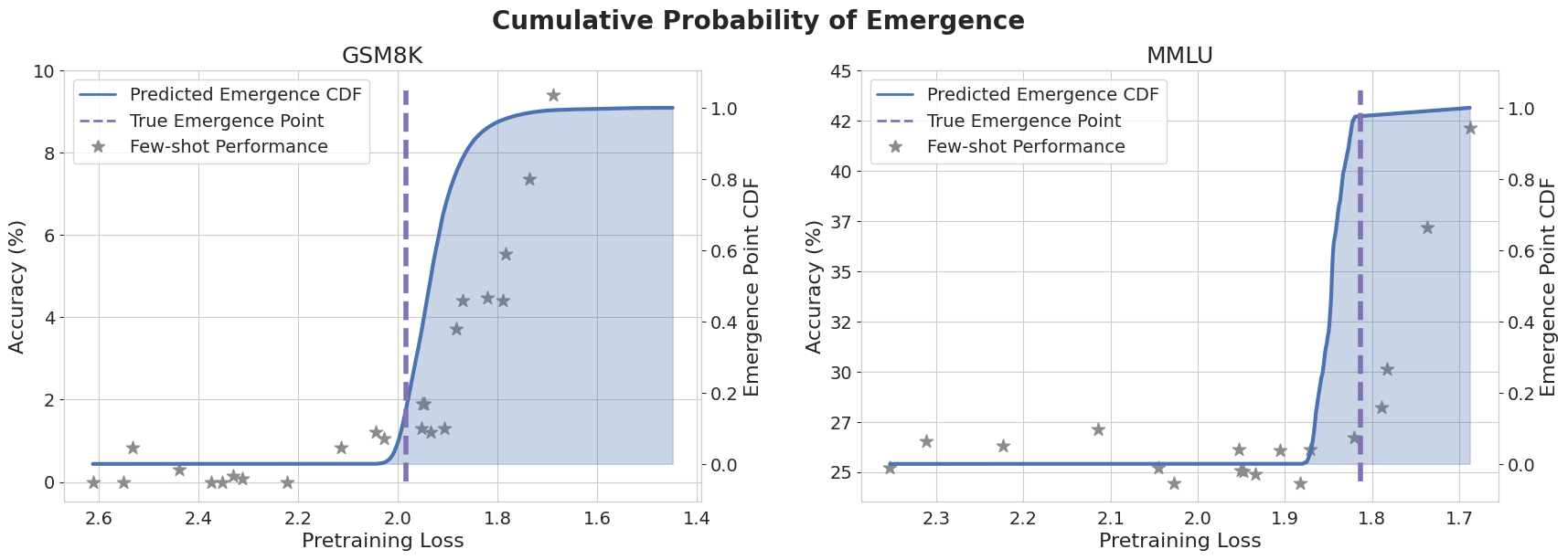}
    \caption{\footnotesize{\textbf{The cumulative distribution function (CDF) of our emergence posterior on GSM8K and MMLU} (see Appendix~\ref{app:full_plots} for all tasks). The CDF's slope peaks at the mode of the distribution. We see that the distribution spikes near the true emergence point, followed by a moderately long tail.}}
    \label{fig:mcmc}
\end{figure}

\begin{figure}
    \centering
    \includegraphics[width=0.99\linewidth]{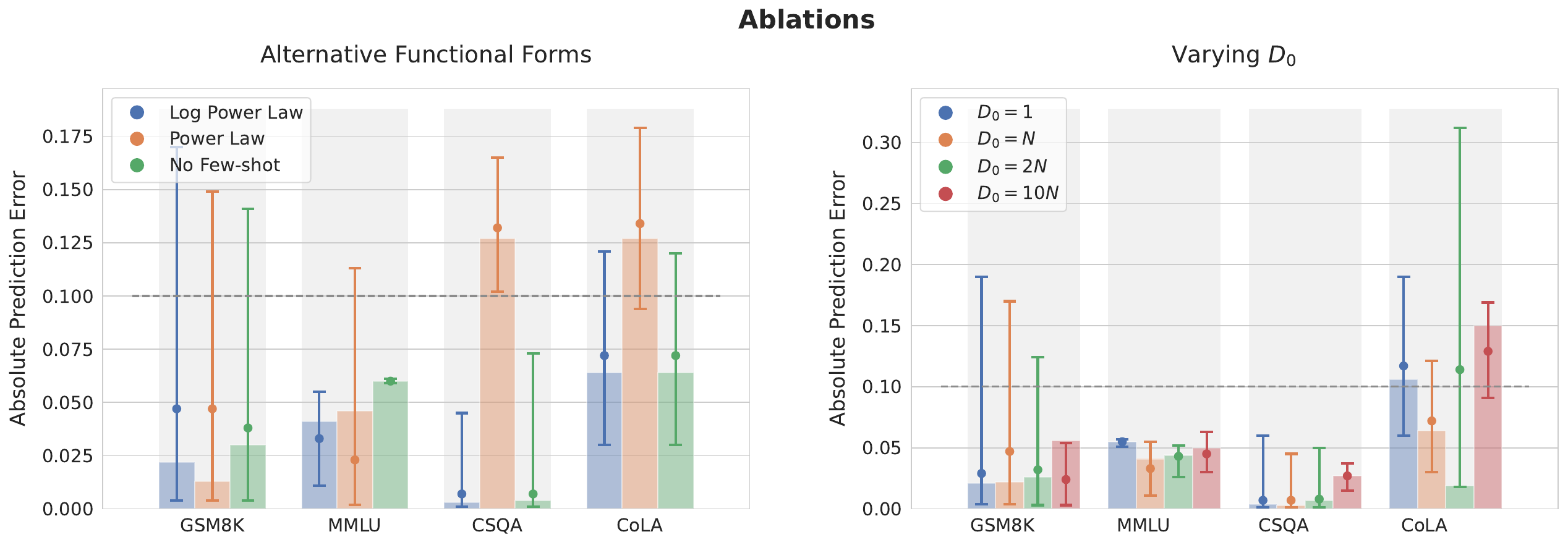}
    \caption{\footnotesize{\textbf{Ablations.} The bar height represents the MLE prediction error (lower is better). The error bar represents the 5th and 95th percentile errors obtained from MCMC posterior sampling, and the circle is the median. \textbf{Left:} \textbf{comparing emergence law functional forms.} ``Log Power Law'' and ``Power Law'' refer to different functional forms for $E_{\theta}(D)$. ``No Few-shot'' is the ``Log Power Law'' without the $\Delta$ parameter. We see that removing the log worsens predictions, and the $\Delta$ has a minimal effect on accuracy. \textbf{Right:} \textbf{varying the low data extrapolation limit $D_0$.} $N$ is the number of few-shot examples. We see that within a reasonable range (e.g., $<10N$) the value of $D_0$ has minimal impact on accuracy.}}
    \label{fig:ablation_bars}
\end{figure}

\paragraph{Models and tasks.} Since fitting our emergence law requires access to a number of intermediate model checkpoints, we use the same 3B, 7B and 13B intermediate OpenLLaMA V1 checkpoints used in Section~\ref{sec:finetuning_interacts}. We conduct experiments using 4 standard NLP benchmarks---MMLU, GSM8K, CommonsenseQA (CSQA), and CoLA---for which we observe emergence in the few-shot setting with the most capable checkpoints. On each of these tasks, none of the 3B checkpoints have emerged, whereas the 7B and 13B models undergo emergence by the end of training. We therefore only use the 3B model checkpoints for fitting our emergence law, and then make use of the 7B and 13B checkpoints for validating the accuracy of our predictions. We include additional details in Appendix~\ref{app:details}.

\paragraph{Fitting the emergence law.} Following the procedure in~\cite{hoffmann2022training}, we obtain an MLE fit for the emergence law, by applying the L-BFGS optimizer. We include additional details in Appendix~\ref{app:fitting}. We estimate a probability distribution for the point of emergence, by taking 100k samples from the No-U-Turn Sampler~\citep{hoffman2014no}.

\paragraph{Evaluation.} We determine the ground-truth point of emergence on each of our tasks, by fitting a ReLU (Equation~\ref{eq:relu}) to the full set of few-shot results from all model checkpoints (3B, 7B, and 13B), using the procedure described in Appendix~\ref{app:fitting}. We report the absolute difference between our MLE prediction and the ground-truth. Using our MCMC probability distribution, we also report a 90\% confidence interval for the prediction error. We consider a prediction successful if our MLE prediction falls within 0.1 nats of the true emergence point.

\subsection{Can our Emergence Law Successfully Predict the Point of Emergence?}
\label{sec:full_fits_main_paper}

\begin{figure}
    \centering
    \includegraphics[width=0.99\textwidth]{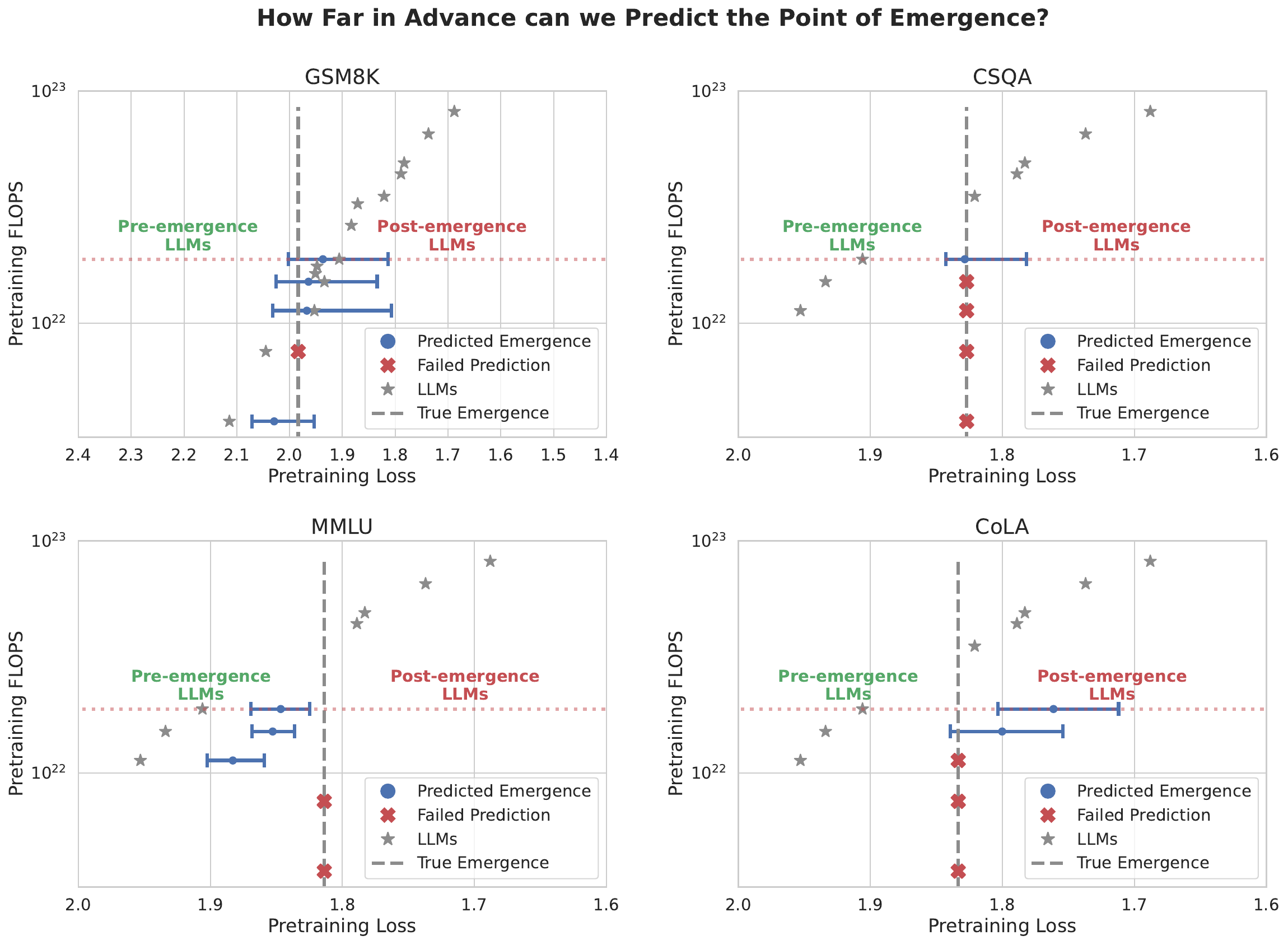}
    \caption[Caption for LOF]{\footnotesize{\textbf{How far in advance can we predict emergence?} We hold out checkpoints to see how far in advance, in pretraining FLOPS, we can successfully predict emergence. The y position of each blue bar corresponds to the FLOPS needed to train the most capable model used for fitting. The blue circle represents the median of the MCMC posterior, and the error bar represents the 5th to 95th percentiles. If the MLE prediction error is $>0.1$ nats, we consider that prediction unsuccessful and denote it with a red-cross\addtocounter{footnote}{-1}\footnotemark. On MMLU we can predict emergence using models trained with $\sim10^{22}$ FLOPS, but no fewer. The earliest post-emergence model on MMLU was trained with $\sim5*10^{22}$ FLOPS, \textbf{hence we can predict 4-5x the FLOPS in advance on MMLU}. On GSM8K we also predict \textbf{4x} the FLOPS in advance\footnotemark. However, on CoLA and CommonsenseQA we only predict \textbf{2x} the FLOPS in advance.}}
    \label{fig:flops}
\end{figure}

In Figure~\ref{fig:fits}, we plot the MLE fit for our emergence law on all four tasks. We see that \textbf{on all tasks, our predictions are very accurate, falling well within 0.1 nats of the true emergence point}\footnote{On GSM8K three of our 3B checkpoints are technically post-emergence according to the ground-truth. However, in absolute terms, these checkpoints have near-trivial performance ($< 3\%$). Therefore without access to larger models these checkpoints would be considered pre-emergence. Additionally, in Figure~\ref{fig:flops}, we see that even in the absence of these checkpoints, we can still make accurate predictions.}. In Figure~\ref{fig:mcmc}, we additionally plot the cumulative distribution function (CDF) of the MCMC emergence posterior on MMLU and GSM8K (see Appendix~\ref{app:full_plots} for all tasks). We see that the distribution generally spikes near the true point of emergence, followed by a moderate sized tail: our emergence law is generally confident about when emergence will occur but has some uncertainty about the possibility of emergence occurring later. Overall, these results serve to validate the predictive accuracy of our emergence law.

\subsection{Comparing Emergence Law Design Decisions}

We would now like to understand how the main decisions in our method effect prediction accuracy. On all four tasks, we ablate: \textbf{1)} the functional form of the emergence law; \textbf{2)} the value of the low data extrapolation limit $D_0$; \textbf{3)} the specific set of model checkpoints and finetuning data amounts selected for fitting the emergence law; and \textbf{4)} the statistical approach used for estimating uncertainty. We detail each ablation below.

\paragraph{Comparing functional forms.}

Recall, we model the emergence shift $E_{\theta}(D)$ using a power law in the log of the amount of finetuning data (Section~\ref{sec:emerg_pred_by_modeling_shift}). This function is then used to parametrize a ReLU with $E_{\theta}(D)$ as the elbow (Section~\ref{sec:fitting_emerg_law}). We finally add an optional $\Delta$ parameter to account for the pre-emergence few-shot data (Section~\ref{sec:full_fewshot_form}). In this section, we ablate both the functional form for $E_{\theta}(D)$ and the choice to include $\Delta$. Specifically, for $E_{\theta}(D)$ we consider a power law in $D$, rather than the log of $D$ (otherwise unchanged). We see in Figure~\ref{fig:ablation_bars} (left) that across all tasks, the log power law generally makes the best predictions. Furthermore, the inclusion of $\Delta$ marginally improves predictions but has an overall small effect. These findings validate the efficacy of our emergence law functional form.

\paragraph{Varying the low data limit.}

\addtocounter{footnote}{-1}
\footnotetext[\thefootnote]{In some cases the failed predictions would be well off the plot and we want to keep the axis bounds constrained for presentation clarity. We include the full results in Appendix~\ref{app:emergence_law_data_ablations}.}
\addtocounter{footnote}{1}
\footnotetext[\thefootnote]{We count the earliest successful prediction for this calculation. However, GSM8K has a failed prediction between two successes, likely due to noise. In Appendix~\ref{app:emergence_law_data_ablations}, we see that this failed prediction is just outside the success threshold, with much of the error bar falling well within 0.1 nats.}

As described in Section~\ref{sec:emerg_pred_by_modeling_shift}, we predict the point of emergence by taking a limit under $E_{\theta}(D)$ into the low data extrapolation limit $D_0$. By default we set $D_0$ to the number of examples in the few-shot prompt $N$. Here, we ablate the effect of choosing different values for $D_0$. We set $D_0$ to a smaller value of $1$ (e.g., the smallest integer for which $\log(D)$ is well defined), and also to larger values of $2N$ and $10N$. In Figure~\ref{fig:ablation_bars} (right), we see that setting $D_0$ to $N$ generally results in the best predictions, but varying $D_0$ within a reasonable range (e.g., $<10N$) has minimal impact on accuracy, demonstrating the robustness of our emergence law to the choice of $D_0$.

\paragraph{Additional ablations.}

In Appendix~\ref{app:emergence_law_data_ablations} we also ablate our procedure for selecting the empirical examples used for fitting the emergence law (Section~\ref{sec:collecting_empirical_data}). We find that, indeed, the smaller finetuning data amounts are often important for making accurate predictions. However, in some cases we find that it is possible to make accurate predictions using far fewer finetuning data amounts and model checkpoints than we used in our main experiments, suggesting that there may be room for future work to greatly improve both the efficiency and effectiveness of our data collection procedure. Finally, in Appendix~\ref{app:uncertainty_ablations}, we compare our MCMC uncertainty intervals with those obtained via bootstrap sampling. We find that in general the two approaches to uncertainty estimation yield similar intervals.

\subsection{How Far in Advance can we Predict the Point of Emergence?}
\label{sec:pred_advance_amount}

\begin{figure}
    \centering
     \includegraphics[width=0.99\textwidth]{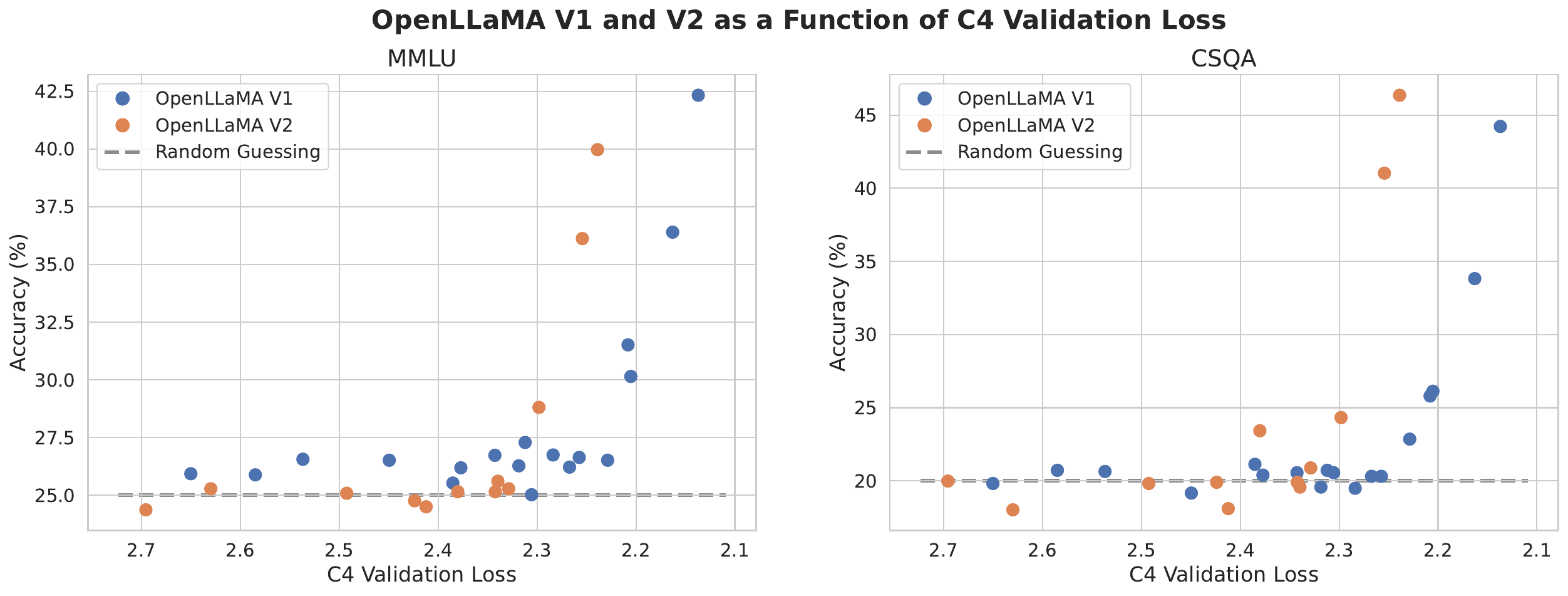}
    \caption{\footnotesize{\textbf{Comparing OpenLLaMA V1 and V2 emergence.} On both MMLU and CommonsenseQA, the V2 models emerge first, suggesting that the V2 pretraining data is likely higher quality.}}
    \label{fig:v1_v_v2}
\end{figure}

We would like to understand how far in advance, in terms of pretraining FLOPS, we can successfully make predictions. To do this, we hold out additional 3B checkpoints when fitting the emergence law. In Figure~\ref{fig:flops}, we plot each emergence prediction against the FLOPS needed to train the most capable model used for fitting. We find that \textbf{the degree to which we can predict emergence in advance is somewhat task dependent.} In particular, on GSM8K and MMLU we are able to reliably make predictions well in advance. On CommonsenseQA and CoLA, on the other hand, we find that our ability to make advance predictions is more limited. Comparing the FLOPS of the earliest point at which we can predict emergence against the FLOPS required for training the earliest post-emergence checkpoint, gives an estimate of how far in advance we can predict. We find that on MMLU and GSM8K we can predict emergence up to \textbf{4.3x and 3.9x FLOPS} in advance respectively. However, on CommonsenseQA and CoLA we are only able predict \textbf{1.9x and 2.3x} in advance respectively.

\section{A Case Study of Real World Uses for Emergence Prediction}
\label{sec:real_world_uses}

Now that we have validated the efficacy of our emergence law approach, we demonstrate two proof-of-concept applications of it: \textbf{1)} cheaply assessing pretraining data quality; and \textbf{2)} predicting the emergence of more complex capabilities, which may only appear in future frontier models. We detail our experiments for both of these settings below.

\subsection{Cheaply Evaluating Pretraining Data Quality}
\label{sec:data_quality}

Emergence prediction can enable model 
developers to more cheaply make modeling decisions on the basis of downstream scaling trends. Without emergence prediction, making such decisions can be costly, requiring the training of large models before seeing any signal. Our emergence law can predict emergence up to 4x the FLOPs in advance, and as such, it presents a potentially useful way to make these decisions more cheaply. Pretraining data quality is one such decision which can be costly to evaluate~\citep{thrush2024improvingpretrainingdatausing}. The downstream capabilities associated with a particular pretraining dataset (e.g., coding) may only emerge at very large compute budgets~\citep{blakeney2024doesdatasparkjoy}, making it expensive to iterate on data. We can use emergence prediction to enable a cheaper data iteration cycle.

\paragraph{Background.} To test the use of emergence prediction for evaluating data quality, we extend our experiments with OpeLLaMA V1 in Section~\ref{sec:evaluating_emergence}, by also experimenting with OpenLLaMA V2~\citep{openlm2023openllama} on MMLU. The V1 and V2 models only differ in that they were trained on different corpa (Appendix~\ref{app:emergence_openllama_v2}). As a result, we should expect their emergence points to differ. The series that emerges earlier should be preferable\footnote{While the earlier emergence could show a more gradual improvement post-emergence, we do not observe this on either task in Figure~\ref{fig:v1_v_v2}. We leave further investigation of this possibility to future work.}.

Since the V1 and V2 models were pretrained on different corpa, their pretraining losses are not comparable. To compare them, we use held-out loss on the C4 validation set as our independent variable $L(\text{M})$. In Figure~\ref{fig:v1_v_v2}, we plot few-shot performance as a function of C4 validation loss for both model series on MMLU and CommonsenseQA. We see that the V2 series emerges earlier than V1, suggesting that the V2 data is higher quality.

\begin{figure}
    \centering
    \includegraphics[width=0.99\textwidth]{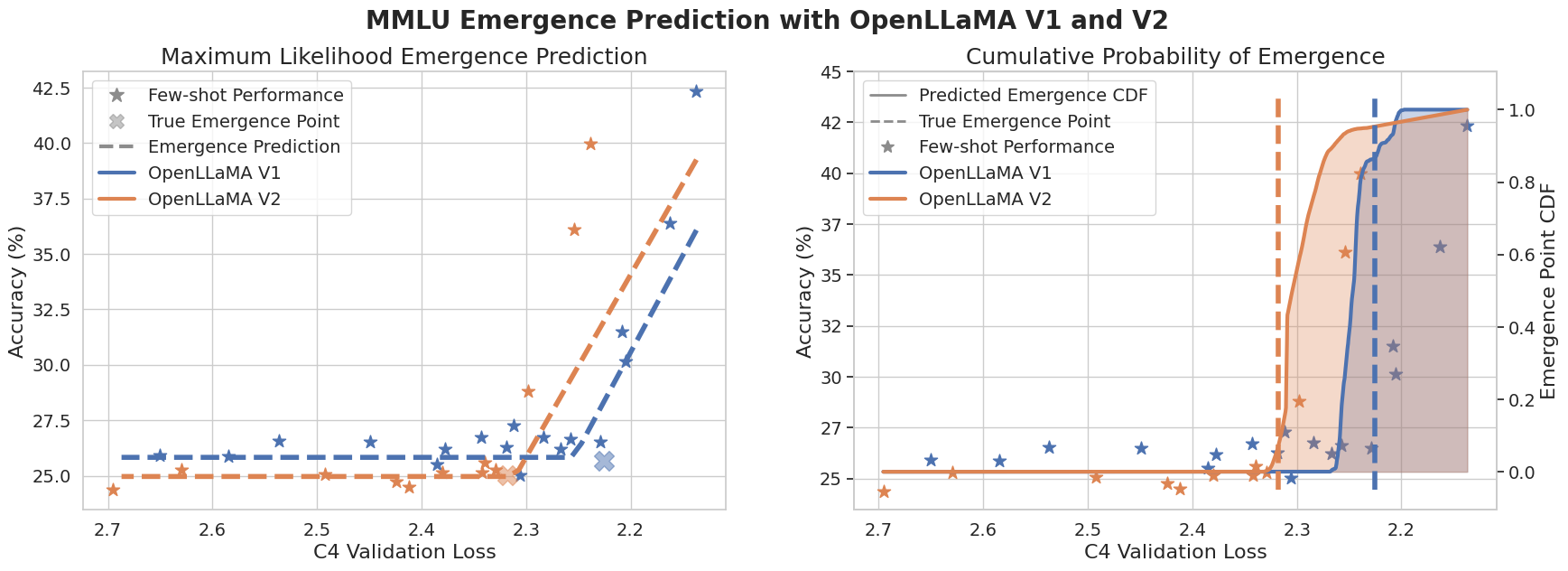}
    \caption{\footnotesize{\textbf{Comparing emergence predictions for OpenLLaMA V1 and V2 on MMLU.} We plot the MLE predictions (left) and the CDFs (right) for both series. While our focus is on predicting the specific point of emergence (e.g., the ReLU elbow), we plot the full ReLU for visual clarity. The V2 models are correctly predicted to emerge before V1, providing initial evidence that our approach can be used to evaluate data quality. See Appendix~\ref{app:full_plots} for plots with all the data used for fitting.}}
    \label{fig:v1_v_v2_pred}
\end{figure}

\paragraph{Predicting data quality.} We now predict the emergence for both series. Similar, to the V1 series, the 3B V2 checkpoints are all pre-emergence. Therefore, just as we did in Section~\ref{sec:evaluating_emergence}, we hold out the larger models and only use the 3B checkpoints for fitting (see Appendix~\ref{app:details} for details). We see in Figure~\ref{fig:v1_v_v2_pred}, that \textbf{our emergence law predicts OpenLLaMA V2 to emerge before V1.} In both cases our prediction error is also well below 0.1 nats, providing initial evidence that emergence laws can be used to cheaply evaluate data quality.

\subsection{Predicting the Capabilities of Future LLMs: a Case Study with LLaMA 2}
\label{sec:future_models}

\begin{figure}
    \centering
    \includegraphics[width=0.99\textwidth]{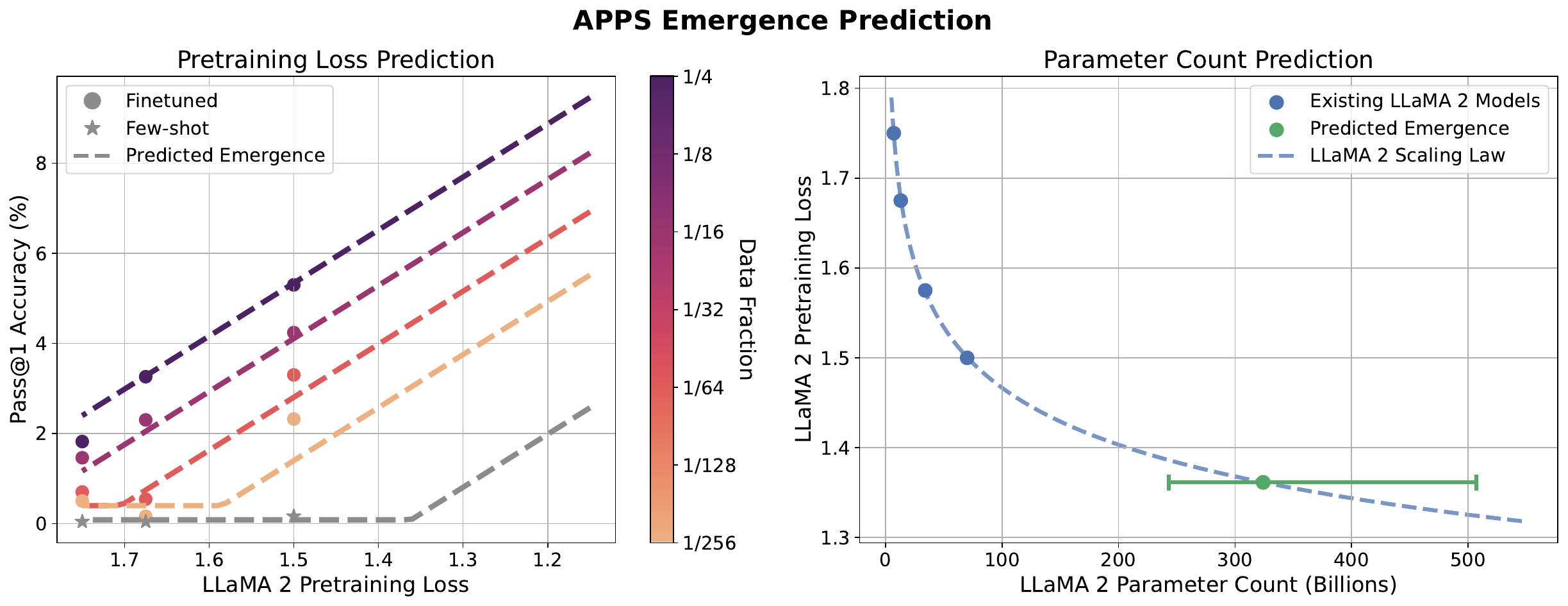}
    \caption{\footnotesize{\textbf{Predicting emergence on APPS with LLaMA 2.} On the left, we plot our MLE prediction. On the right, we convert this loss-based prediction into parameter count under the LLaMA 2 scaling law. The green point represents the MLE prediction, and the error bar represents the 5th to 95th percentiles under the MCMC posterior. We predict that emergence would most likely occur at $\sim325$B parameters with a wide error bar from $\sim250$B to $\sim500$B parameters. For visual clarity, the left plot includes a subset of the full data used for fitting (see Appendix~\ref{app:full_plots} for all).}}
    \label{fig:apps_params}
\end{figure}

Emergence prediction can potentially be used to predict frontier LLM capabilities, including those which are safety relevant. However, these capabilities will be much more complex than the tasks studied in Section~\ref{sec:evaluating_emergence}, all of which emerge well before today's frontier. To therefore validate the efficacy of our emergence law on more complex tasks, we conduct a proof-of-concept experiment using LLaMA 2 on the difficult APPS coding benchmark.

\paragraph{Background.} APPS is much more challenging than the tasks studied in Section~\ref{sec:evaluating_emergence}. Namely, all LLaMA 2 models demonstrate approximately random greedy pass@1 accuracy ($0.04\%$ for 7B and 13B and $0.16\%$ for 70B), making all of these models pre-emergence. We note that LLaMA 2 potentially under-performs on coding tasks because the pretraining data was reportedly not optimized for code~\citep{rozière2024codellamaopenfoundation,patel2024zuckerberg}. Nonetheless, on easier code benchmarks like human-eval~\citep{chen2021evaluatinglargelanguagemodels} and MBPP~\citep{austin2021programsynthesislargelanguage}, LLaMA 2 achieves non-trivial performance~\citep{touvron2023llama2openfoundation,rozière2024codellamaopenfoundation}, suggesting that some code was still present in pretraining. We should therefore expect that if Meta were to scale beyond 70B on the same data, we would eventually see emergence on APPS. We aim to predict this point of emergence. While we cannot verify the accuracy of this prediction, this experiment will show that the general trend in which finetuning shifts the point of emergence (Section~\ref{sec:finetuning_interacts}) still holds for more complex tasks and with larger LLMs, acting as a proof-of-concept for forecasting emergence closer to the frontier.

\paragraph{Predicting emergence on APPS.} To fit an emergence law on APPS, we finetune all three open LLaMA 2 models --- 7B, 13B, and 70B --- on subsets of the APPS training split (Appendix~\ref{app:details} for details). We evaluate greedy decoding on the test set. In Figure~\ref{fig:apps_params} (left), we see that our observations from Section~\ref{sec:finetuning_interacts} --- finetuning shifts the point of emergence --- also hold on APPS, suggesting that our method can transfer to more complex tasks. We predict the point of emergence to be $\sim$0.15 nats beyond 70B LLaMA 2. We convert this prediction into parameter count, by mapping it onto the LLaMA 2 scaling law (see Appendix~\ref{app:llama2_scaling_law_details}). After this transformation, we find that \textbf{emergence would most likely occur at $\sim$325B parameters} with a wide error bar between $\sim$250B and $\sim$500B parameters (Figure~\ref{fig:apps_params} right). While we cannot validate our prediction accuracy, this experiment serves as a proof-of-concept that our approach can be used to predict capabilities closer to the frontier.

\section{Limitations and Future Directions}
\label{sec:discussion}

In Section~\ref{sec:pred_advance_amount}, we found that our specific emergence prediction approach (e.g., emergence law) can accurately predict the point of emergence up to 4x the FLOPS in advance, representing meaningful progress on the challenging unsolved problem of emergence prediction. However, these results fall far short of the 1000x demonstrated in~\cite{openai2024gpt4} for predicting post-emergence downstream capabilities. Such advance predictions may be necessary in particularly high-stakes settings; therefore further progress is needed. Below we outline the limitations of our approach and discuss possible future directions.

\paragraph{Better data selection may improve predictions.} Our data collection procedure was not specifically designed to maximize the degree to which we can predict emergence in advance and thus can likely be improved. For example, some datapoints are more important than others for making accurate predictions (see Appendix~\ref{app:emergence_law_data_ablations}). It may therefore be possible to improve predictions by borrowing ideas from the active learning literature~\citep{settles2009active} to select the maximally informative datapoints for emergence law fitting.

\paragraph{Why does finetuning result in an emergence shift?} We have a limited understanding of \emph{why} finetuning shifts the point of emergence. In Appendix~\ref{app:peft}, we experimented with alternative methods for inducing an emergence shift, finding model parameter updates to be an important factor. However, these results do not present a complete explanation: we do not understand how finetuning interacts with emergence at a mechanistic level. Is the role of finetuning to accelerate an underlying phase change inside the model, or is it just surfacing existent latent capabilities? This is an exciting question for future work to explore.

\paragraph{Predictions may not transfer in all settings.} All of our experiments are conducted using transformer checkpoints, which only meaningfully difer in the amount of pretraining data used and the parameter count. It is not well understood whether LLMs with substantially different architectures (e.g., state-space models)~\citep{tay2022scaling,arora2023zoology,gu2023mamba} or trained differently (e.g., using distillation) will demonstrate the same downstream capabilities at a given pretraining loss. It is therefore possible that our emergence laws will not transfer in these settings. Future work work should investigate this possibility.

\paragraph{Task-specific finetuning may be limited.} Finetuning models broadly on many tasks at once with a limited amount of data, can often fail to substantially improve a model's general capabilities~\citep{gudibande2023false}. While we instead focused on finetuning for specific tasks in this work, it is possible that in certain settings (e.g., LM agents), models may be required to compose many different skills together. In this case, finetuning may be less effective. While we observed encouraging results on the challenging APPS benchmark in Section~\ref{sec:future_models}, further evaluation of the limits of our approach is needed.

\section*{Acknowledgements}

We thank Ruiqi Zhong, Nick Lourie, Nicholas Tomlin, Jason Wei, Kevin Liu, Jonathan Uesato, Young Geng, Daniel Bauman, and Jiayi Pan for discussion and feedback on earlier drafts of our paper. Charlie Snell is supported by the OpenAI Superalignment Fellowship. This research was supported with Cloud TPUs from Google’s TPU Research Cloud (TRC).

\bibliography{colm2024_conference}
\bibliographystyle{colm2024_conference}

\appendix
\section{Appendix}

\subsection{\textbf{Additional Related Work}}

\paragraph{Neural scaling laws.} A number of works have studied the scaling behavior of language model pretraining loss~\cite{kaplan2020scaling,hoffmann2022training,aghajanyan2023scaling,muennighoff2024scaling,henighan2020scaling,krajewski2024scaling}. Most related to our work are~\cite{hernandez2021scaling} and~\cite{isik2024scaling}, which study how finetuned language models scale as a function of model size and finetuning data amount. However, these works critically differ from ours, in that they study performance on settings which are already smoothly improving with scale.

\paragraph{Scaling laws for downstream metrics.} In addition to understanding the scaling behavior of upstream metrics, a number of works have proposed methods for modeling the scaling behavior of downstream performance~\citep{ivgi-etal-2022-scaling,caballero2023broken,gadre2024language,isik2024scaling,owen2024predictablelanguagemodelbenchmark,ruan2024observationalscalinglawspredictability,hu2024predictingemergentabilitiesinfinite}. All of these works differ from our setting in that they assume the downstream metric of interest is already showing signs of smooth improvement as a function of model scale (i.e., post-emergence). Most related to our work is concurrent work from~\cite{blakeney2024doesdatasparkjoy}, which shows that by up-sampling domain-specific data at the end of pretraining, downstream LLM performance on certain tasks can be boosted, even on emergent tasks. Our work instead focuses on using the boost from finetuning to help predict the point of emergence in the few-shot setting.

\subsection{\textbf{Alternative Methods for Shifting the Point of Emergence}}
\label{app:peft}

\begin{figure*}
    \centering
    \subfigure{
        \includegraphics[width=0.48\textwidth]{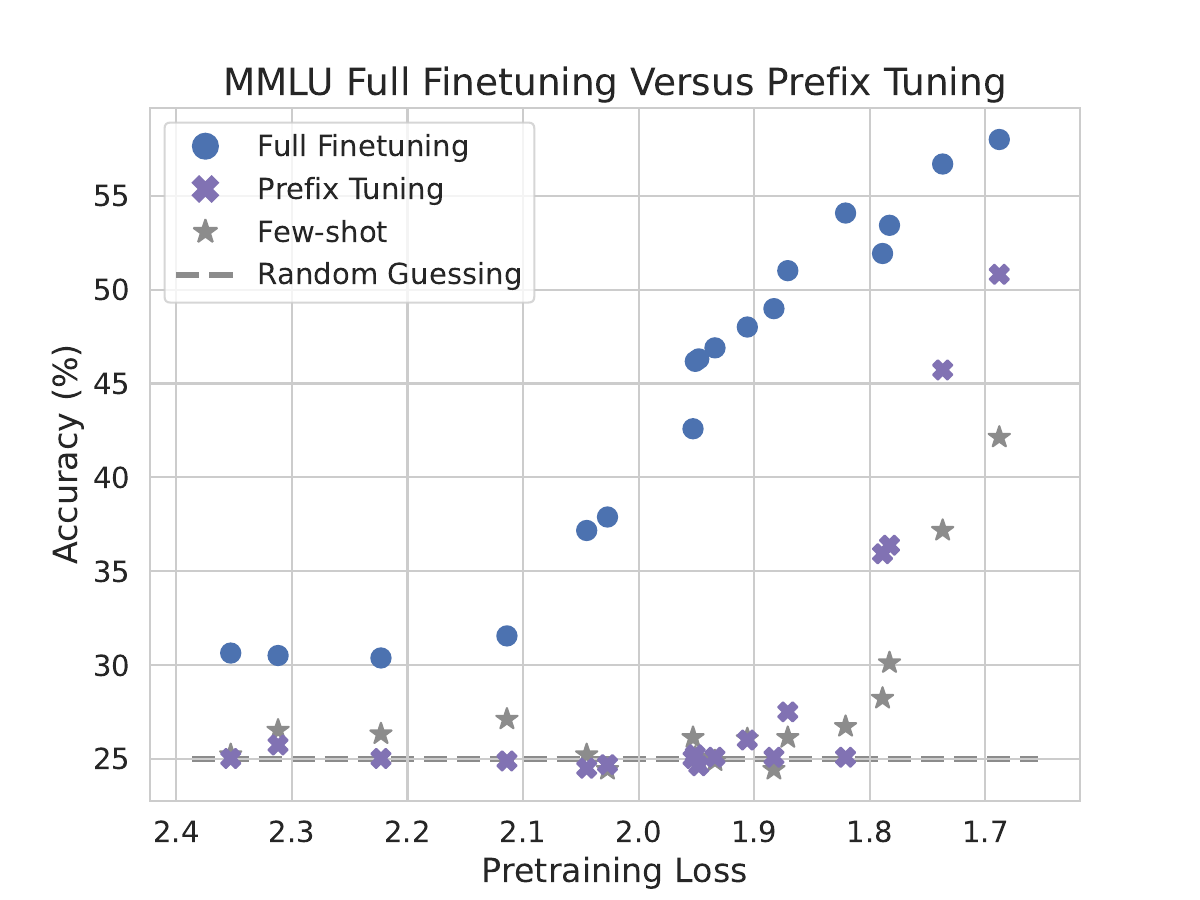}
        \label{fig:prefix_mmlu}
    }\hfill
    \subfigure{
        \includegraphics[width=0.48\textwidth]{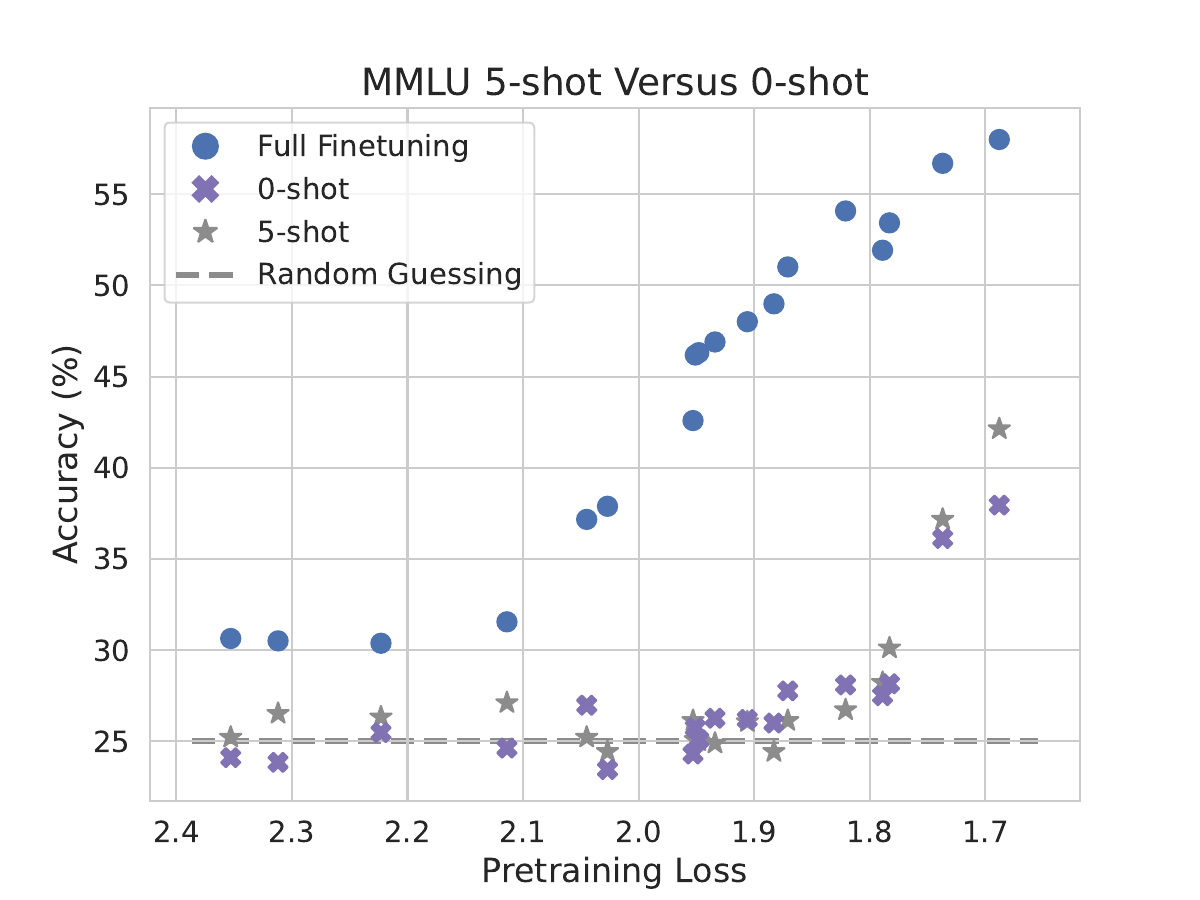}
        \label{fig:zeroshot_mmlu}
    }\\
    \caption{\footnotesize{One the left we compare full fine-tuning against continuous prefix tuning on MMLU. We find that prefix tuning provides effectively no shift to the point of emergence, despite improving the performance of post-emergence models. On the right we compare 0-shot verses 5-shot prompting on MMLU. We see that using fewer shots has no meaningful effect on the point of emergence. Together these results suggest that the ability for prompt tuning to shift the point of emergence is very limited.}}
    \label{fig:prefix_tuning_zero_shot}
\end{figure*}

\begin{figure*}[h]
    \centering
    \includegraphics[width=0.99\textwidth]{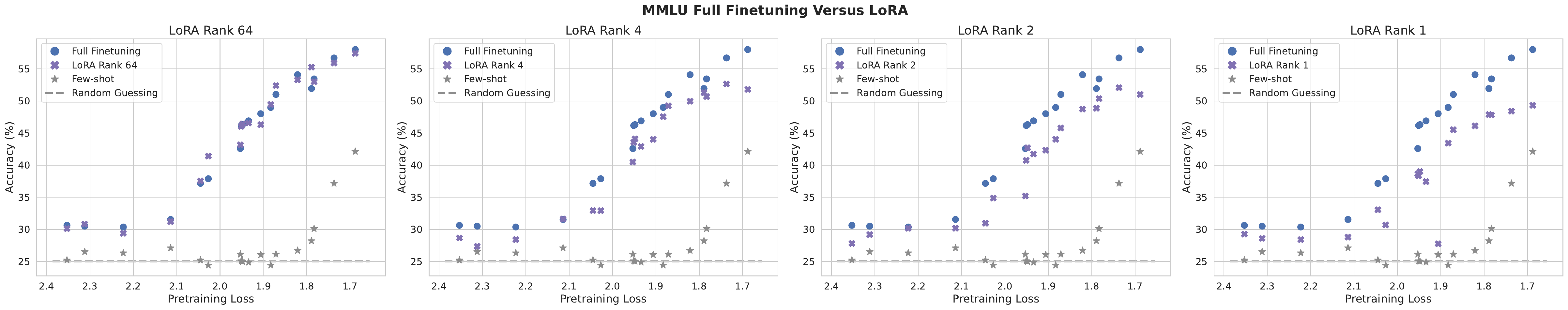}
    \caption{\footnotesize{Comparing LoRA finetuning, with rank 1, 2, 4, and 64 against full finetuning on MMLU. We see that LoRA finetuning even with rank 1 shifts the point of emergence to a comparable degree to that of full finetuning.}}
    \label{fig:lora}
\end{figure*}

We are interested in understanding what other methods can induce a shift in the point of emergence. We therefore conduct additional experiments on MMLU with 1) varying the number of shots in the prompt, 2) performing continuous prefix tuning~\cite{li2021prefix}, and 3) using low-rank finetuning with LoRA. We see in Figure~\ref{fig:prefix_tuning_zero_shot} that both prefix tuning and using fewer shots in the few-shot prompt have little effect on shifting the point of emergence. On the other hand, in Figure~\ref{fig:lora} we see that low-rank finetuning shifts the point of emergence to a comparable degree to that of full finetuning, even in the rank-1 setting.

Together these results suggest that updating the model's parameters may be necessary for shifting the point of emergence. Though further exploration of this phenomenon is needed. In particular, it is possible that using hundreds or thousands of shots in the prompt may enable more of a shift than what we observed~\citep{agarwal2024manyshotincontextlearning}.

In each of these experiments, we use all 3B, 7B, and 13B OpenLLaMA V1 model checkpoints. For the prefix tuning baseline we use a prefix length of 8, parameterized by a 2-layer MLP with input and hidden-dim 512. We train with a learning rate of 3e-4 and keep all other hyperparameters the same as our full finetuning experiments. We found attempts to increase the capacity of the prefix tuning (e.g., using a longer prefix or dropping the MLP and directly tuning the embeddings) to make training more unstable and generally yield worse performance. For LoRA we use the same finetuning hyper-parameters as full fine-tuning, including the learning rate. To ensure that the LoRA updates are of similar magnitude to full finetuning with the same learning rate, we set the LoRA $\alpha$ hyper-parameter to be equal to the model's hidden dimension. We found this setup to yield the best results with minimal changes between our full finetuning and LoRA setup.

\subsection{\textbf{Emergence with Continuous Metrics}}

\begin{figure}
    \centering
    \subfigure{
        \includegraphics[width=0.48\textwidth]{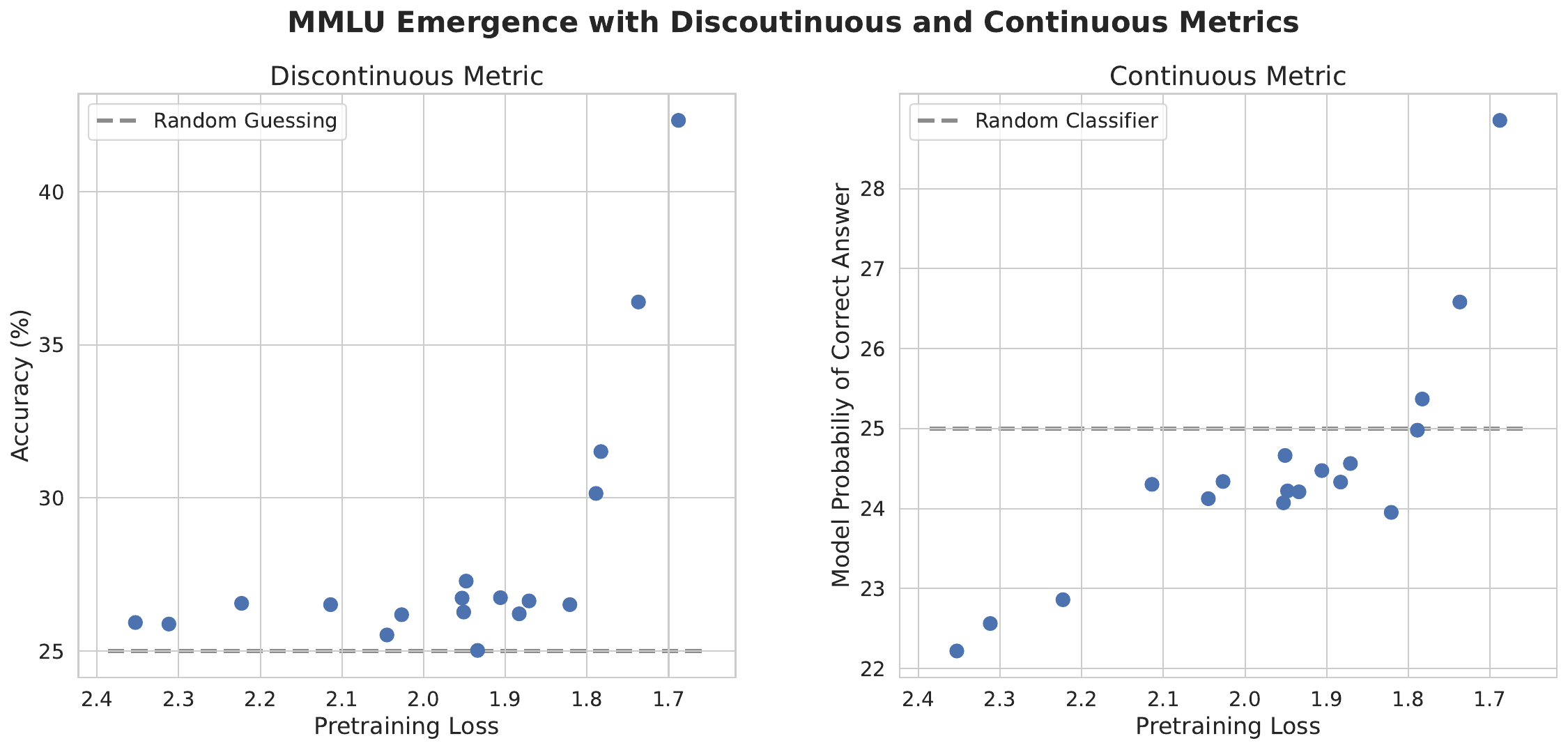}
        \label{fig:nonsmooth_emergence_mmlu}
    }\hfill
    \subfigure{
        \includegraphics[width=0.48\textwidth]{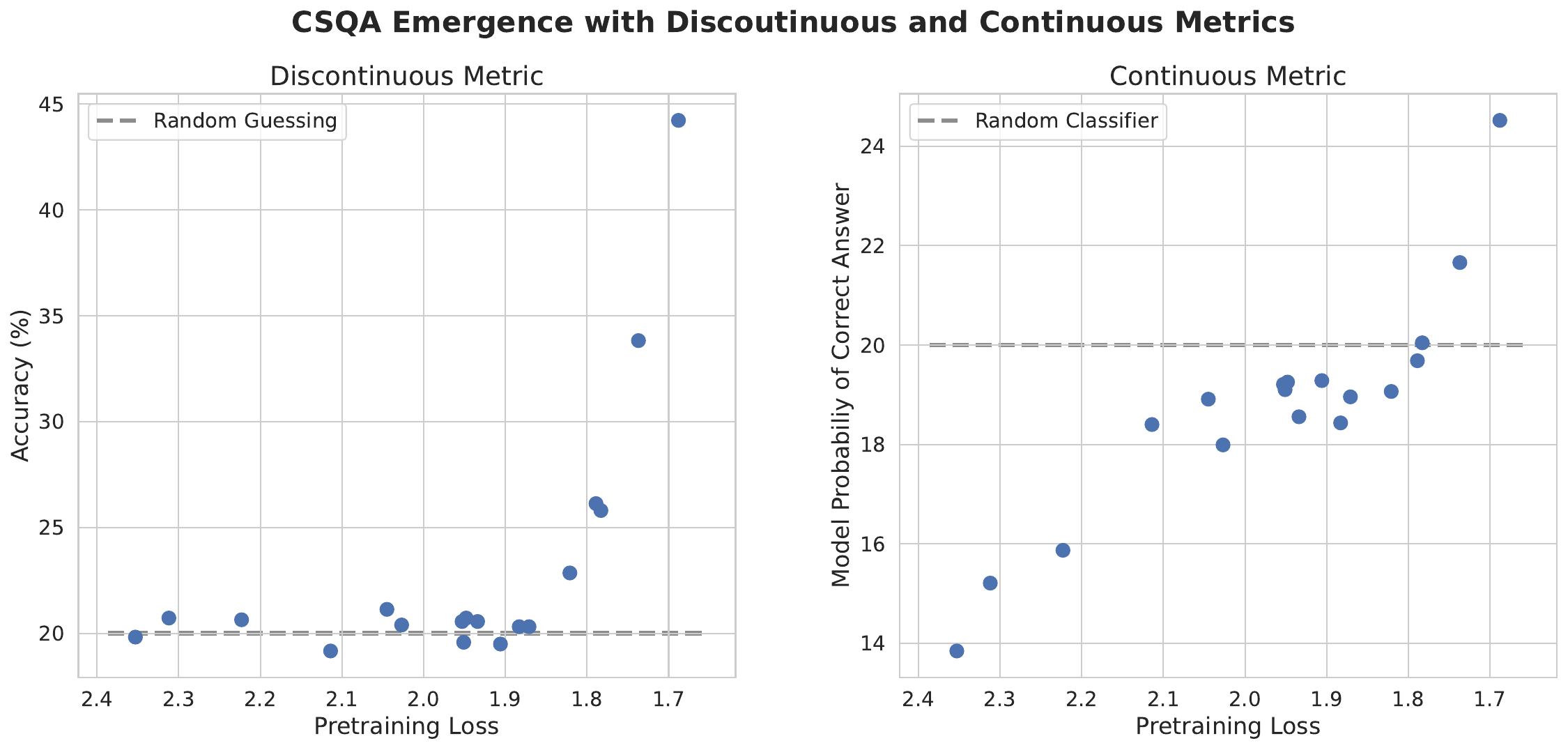}
        \label{fig:nonsmooth_emergence_csqa}
    }
    \caption{\footnotesize{On a standard 5-shot MMLU and 6-shot CommonsenseQA (CSQA) evaluation, we observe emergence using both the standard correct answer accuracy evaluation and a continuous LLM log-probability metric.}}
    \label{fig:nonsmooth_emergence}
\end{figure}

In Figure~\ref{fig:nonsmooth_emergence} we present two LLM tasks (MMLU and CommonsenseQA) in which we observe emergence with both continuous and discontinuous metrics.

\subsection{\textbf{Finetuning and Few-shot Emergence Across Model Sizes}}
\label{app:finetuning_fewshot_across_sizes}

\begin{figure}
    \centering
    \includegraphics[width=0.99\textwidth]{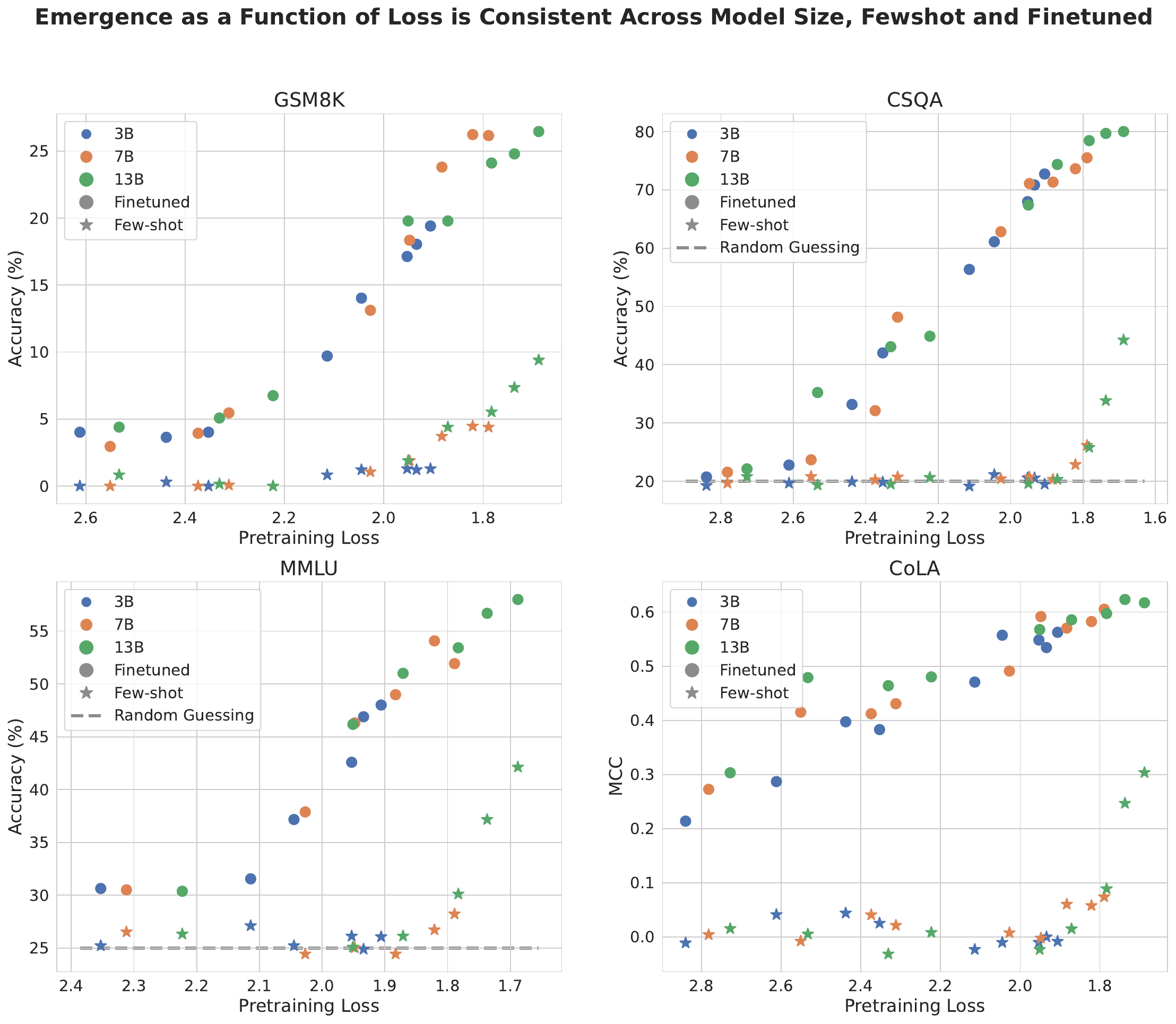}
    \caption{\footnotesize{We plot few-shot and full data finetuning performance as a function of pretraining loss using all 3B, 7B, and 13B model checkpoints for all tasks. We see that both the point of emergence and the downstream performance scaling thereafter, as a function of pretraining loss, is consistent across model size in both the few-shot and finetuned setting.}}
    \label{fig:across_model_size}
\end{figure}

In Figure~\ref{fig:across_model_size}, we plot emergence as a function of pretraining loss using all 3B, 7B, and 13B OpenLLaMA V1 model intermediate checkpoints in both the few-shot and finetuned setting. We finetune on the full data amount. The specific finetuning hyperparameters and data splits used are detailed in Appendix~\ref{app:details}. For the few-shot results, we use the prompts detailed in Appendix~\ref{app:details}. On all tasks, we see that both the point of emergence and the downstream performance scaling thereafter, as a function of pretraining loss, is consistent across model size in both the few-shot and finetuned setting. In the case of CoLA, finetuning on the full data amount significantly shifts the point of emergence to such a degree that it precedes all model checkpoints that we consider. We see in Section~\ref{app:full_plots} that when using smaller data amounts, the emergence elbow on CoLA is shifted towards stronger models and is thus visible with our checkpoints.

\subsection{\textbf{Emergence Law Fitting}}
\label{app:fitting}

For obtaining our emergence law maximum-likelihood fits we follow the procedure in~\cite{hoffmann2022training} and use the L-BFGS optimizer, selecting the best fit from a sweep over initializations. In particular, we first perform a brute-force grid-search over all the values in: $A\in\{0.0, 0.1, 0.2, ..., 4.0\}$, $B\in\{0.0, 0.1, 0.2, ..., 1.0\}$, $k\in\{0.0, 0.05, 0.1, ..., 1.0\}$, $\alpha\in\{1.0, 1.5, 2.0, ..., 10.0\}$, $C\in\{0.0, 0.5, 1.0, ..., 10.0\}$. We select the 100K parameters from this grid-search with the lowest loss and then run the L-BFGS optimizer initialized from each one. We then select the best fit final fit from all optimization runs. We fit the ReLU for the true emergence point using the same procedure.

\subsection{\textbf{Emergence Law Data Ablations}}
\label{app:emergence_law_data_ablations}

\begin{table}
    \centering
    \scriptsize
    \setlength{\tabcolsep}{4pt}
    \setlength{\lightrulewidth}{0.01em}  
    \setlength{\heavyrulewidth}{0.175em}  
    \begin{tabular}{l|rrrr}
        \toprule
        \textbf{Ablation Setting} & \textbf{GSM8K} & \textbf{MMLU} & \textbf{CSQA} & \textbf{CoLA} \\
        \midrule[\heavyrulewidth]
        \addlinespace[0.5em]
        Full Data & \textcolor{green!50!black}{0.022 [0.004, 0.170]} & \textcolor{green!50!black}{0.041 [0.011, 0.055]} & \textcolor{green!50!black}{0.003 [0.001, 0.045]} & \textcolor{green!50!black}{0.064 [0.030, 0.121]} \\
        \midrule[\heavyrulewidth]
        \addlinespace[0.5em]
        -1 Smallest Subset & \textcolor{green!50!black}{0.014 [0.002, 0.051]} & \textcolor{green!50!black}{0.022 [0.010, 0.031]} & \textcolor{green!50!black}{0.051 [0.037, 0.084]} & \textcolor{green!50!black}{0.071 [0.024, 0.097]} \\
        -2 Smallest Subset & \textcolor{green!50!black}{0.047 [0.018, 0.063]} & \textcolor{green!50!black}{0.025 [0.003, 0.032]} & \textcolor{green!50!black}{0.087 [0.063, 0.129]} & \textcolor{red}{0.634 [0.577, 0.711]} \\
        -3 Smallest Subset & \textcolor{green!50!black}{0.005 [0.002, 0.050]} & \textcolor{green!50!black}{0.001 [0.002, 0.025]} & \textcolor{red}{0.988 [0.913, 1.099]} & \textcolor{red}{1.513 [1.291, 1.698]} \\
        \midrule[\lightrulewidth]
        -1 Largest Subset & \textcolor{green!50!black}{0.022 [0.002, 0.032]} & \textcolor{green!50!black}{0.034 [0.024, 0.034]} & \textcolor{green!50!black}{0.045 [0.003, 0.096]} & \textcolor{green!50!black}{0.036 [0.007, 0.090]} \\
        -2 Largest Subset & \textcolor{green!50!black}{0.005 [0.002, 0.054]} & \textcolor{red}{1.017 [1.482, 1.985]} & \textcolor{green!50!black}{0.057 [0.056, 0.059]} & \textcolor{green!50!black}{0.004 [0.002, 0.058]} \\
        -3 Largest Subset & \textcolor{green!50!black}{0.016 [0.002, 0.083]} & \textcolor{red}{0.332 [1.371, 4.200]} & \textcolor{green!50!black}{0.089 [0.077, 0.098]} & \textcolor{green!50!black}{0.044 [0.019, 0.143]} \\
        \midrule[\lightrulewidth]
        Only Subset Sample 1 & \textcolor{green!50!black}{0.006 [0.001, 0.031]} & \textcolor{red}{0.244 [0.223, 0.489]} & \textcolor{red}{1.557 [1.616, 2.573]} & \textcolor{red}{0.179 [0.125, 0.269]} \\
        Only Subset Sample 2 & \textcolor{green!50!black}{0.026 [0.002, 0.067]} & \textcolor{green!50!black}{0.073 [0.059, 0.077]} & \textcolor{green!50!black}{0.035 [0.004, 0.102]} & \textcolor{green!50!black}{0.034 [0.022, 0.047]} \\
        \midrule[\heavyrulewidth]
        \addlinespace[0.5em]
        Last 6 Checkpoints & \textcolor{green!50!black}{0.010 [0.003, 0.113]} & \textcolor{green!50!black}{0.041 [0.007, 0.049]} & \textcolor{green!50!black}{0.075 [0.063, 0.301]} & \textcolor{red}{0.118 [0.087, 0.167]} \\
        Last 5 Checkpoints & \textcolor{green!50!black}{0.047 [0.048, 0.176]} & \textcolor{green!50!black}{0.032 [0.020, 0.038]} & \textcolor{red}{1.165 [1.070, 1.728]} & \textcolor{red}{0.130 [0.099, 0.170]} \\
        Last 4 Checkpoints & \textcolor{green!50!black}{0.080 [0.072, 4.925]} & \textcolor{green!50!black}{0.030 [0.001, 0.042]} & \textcolor{red}{1.734 [1.555, 2.308]} & \textcolor{green!50!black}{0.076 [0.052, 0.111]} \\
        Last 3 Checkpoints & \textcolor{red}{0.159 [0.124, 0.703]} & \textcolor{green!50!black}{0.013 [0.002, 0.059]} & \textcolor{red}{0.986 [0.802, 1.249]} & \textcolor{green!50!black}{0.039 [0.004, 0.077]} \\
        Last 6 Checkpoints, Every Other Even & \textcolor{green!50!black}{0.019 [0.003, 0.162]} & \textcolor{green!50!black}{0.070 [0.059, 0.075]} & \textcolor{red}{1.663 [1.667, 1.781]} & \textcolor{green!50!black}{0.033 [0.007, 0.050]} \\
        Last 6 Checkpoints, Every Other Odd & \textcolor{green!50!black}{0.044 [0.041, 0.126]} & \textcolor{green!50!black}{0.040 [0.024, 0.045]} & \textcolor{green!50!black}{0.037 [0.031, 0.191]} & \textcolor{red}{0.224 [0.190, 0.287]} \\
        \midrule[\lightrulewidth]
        -1 Last Checkpoints & \textcolor{green!50!black}{0.069 [0.003, 0.149]} & \textcolor{green!50!black}{0.043 [0.023, 0.055]} & \textcolor{red}{0.985 [0.858, 1.800]} & \textcolor{green!50!black}{0.026 [0.004, 0.079]} \\
        -2 Last Checkpoints & \textcolor{green!50!black}{0.087 [0.003, 0.176]} & \textcolor{green!50!black}{0.076 [0.046, 0.089]} & \textcolor{red}{0.102 [0.098, 0.104]} & \textcolor{red}{0.165 [0.098, 0.242]} \\
        -3 Last Checkpoints & \textcolor{red}{0.110 [0.010, 0.468]} & \textcolor{red}{0.664 [0.500, 0.959]} & \textcolor{red}{0.616 [0.510, 0.822]} & \textcolor{red}{2.217 [2.031, 2.407]} \\
        -4 Last Checkpoints & \textcolor{green!50!black}{0.044 [0.005, 0.089]} & \textcolor{red}{2.308 [2.347, 57.068]} & \textcolor{red}{0.581 [0.546, 1.169]} & \textcolor{red}{1.039 [0.905, 1.298]} \\
        \bottomrule
    \end{tabular}
    \caption{\footnotesize{Ablating the effect of holding out different finetuning subsets and model checkpoints when fitting the emergence law. We present the absolute error between the maximum likelihood predicted point of emergence and the ground-truth. In brackets we include the 5th and 95th percentile of prediction errors produced by our MCMC posterior sampling. We consider fits where the maximum likelihood prediction is greater than 0.1 nats from the ground-truth to be failures and highlight these cases in red; otherwise we highlight in green. In the top row we present results for the fit obtained using all finetuning data amounts and model checkpoints. In the middle rows (e.g., ``-1 Smallest Subset'' to ``Only Subset Sample 2''), we present ablations in which we hold out various finetuning data subsets, so as to understand the effect of our data subset selection methodology on our predictions. Finally, in the bottom rows, we present ablations in which we hold out various model checkpoints, so as to understand how many checkpoints are needed to obtain good predictions (e.g., ``Last 6 Checkpoints'' to ``-4 Last Checkpoint''). We describe each ablation in more detail in Appendix~\ref{app:emergence_law_data_ablations}.}}
    \label{tab:emergence_law_data_ablations}
\end{table}

We would like to understand what aspects of our emergence law data collection methodology are important for enabling effective emergence predictions. In particular, our data collection involves finetuning several model checkpoints on different subsets of our full finetuning data. In general, we should expect using more model checkpoints and more finetuning data subsets for fitting the emergence law to strictly improve our predictions. However, in practice we may have a limited finetuning budget or may have access to a limited set of model checkpoints. For this reason, we would like to better understand the limits and best practices for selecting the empirical datapoints for emergence law fitting.

Concretely, there are two variables in our emergence law data collection procedure: the set of checkpoints selected, and the set of data subsets we finetune on. In Table~\ref{tab:emergence_law_data_ablations}, we conduct a series of ablations in which we modify both of these variables to understand how they effect our emergence predictions. We discuss each of these ablations below.

\paragraph{Holding out finetuning data amounts.} As described in Appendix~\ref{app:details}, we select the sizes of the finetuning data subsets for fitting the emergence law by first conducting a logarithmic sweep over various data amounts (e.g., $\frac{1}{2}$ the data, $\frac{1}{4}$ the data, $\frac{1}{8}$ the data, etc...). We then then collect additional subsets nearer to the limit at which emergence is visible with our 3B checkpoints. We also finetune on two different sampled subsets at each data amount to account for noise in the subsampling process.

To better understand how our data subset selection procedure effects the accuracy of our emergence predictions, we experiment with holding out the N largest and N smallest subsets when conducting our fits (see the rows ``-N Smallest Subset'' and ``-N Largest Subset'' in Table~\ref{tab:emergence_law_data_ablations}). We see that, with the exception of MMLU, our predicted fits remain fairly accurate when holding out the larger subsets, whereas the holding out smaller subsets can have a larger effect on prediction accuracy. This finding aligns with our intuition that the smaller subsets, which are closer to the low data extrapolation limit which we use to make our emergence predictions, are more critical for making accurate predictions.

We also ablate the effect of using two independent subsamples at each finetuning data amount to account for noise, by fitting our emergence predictions using only 1 subsample. This ablation is presented in the ``Only Subset Sample 1'' and ``Only Subset Sample 2'' rows of Table~\ref{tab:emergence_law_data_ablations}, where each row corresponds to using only one of the two sub-samples for fitting the emergence law. We see that the noise from using only a single subset can result in worse predictions. Therefore including multiple subsamples can be a useful way to integrate out noise when fitting an emergence law.

\paragraph{Holding out model checkpoints.} For each task, we use results from at least 6 and up to 9 intermediate model checkpoints to fit our emergence law, which are mostly evenly spaced (see Appendix~\ref{app:details} for the specific details). We would like to better understand just how many checkpoints are needed to make accurate predictions. Since we model emergence with a ReLU, we will need at least 3 checkpoints, but, in practice, it is likely that more checkpoints will be necessary.

We expect that the later checkpoints, which are closer to the point of emergence, should be more informative for making emergence predictions. Therefore, to understand the minimum number of checkpoints needed to make accurate emergence predictions, we fit our emergence law using only the last N checkpoints (see the ``Last N Checkpoints'' rows in Table~\ref{tab:emergence_law_data_ablations}). We also consider using fewer checkpoints but spacing them out more. Specifically, in the ``Last 6 Checkpoints, Every Other Even'' and ``Last 6 Checkpoints, Every Other Odd'' rows in Table~\ref{tab:emergence_law_data_ablations}, we consider every other checkpoint from the most recent 6 checkpoints, leaving 3 total checkpoints. The odd case represents every other checkpoint shifted by one from the even case. We see that removing checkpoints can cause predictions to worsen, however if the checkpoints are well spaced, as in the ``Every Other'' case, predictions can, in some cases, remain highly accurate with only 3 checkpoints.

Finally in the ``-N Last Checkpoints'' rows in Table~\ref{tab:emergence_law_data_ablations}, we hold out the N final checkpoints. These represent the complete results for the ablations in Section~\ref{sec:pred_advance_amount} aimed at understanding how far in advance our emergence law can make predictions.

\subsection{\textbf{Comparing Uncertainty Estimation Approaches}}
\label{app:uncertainty_ablations}

In Table~\ref{tab:uncertainty_ablation_results}, we compare MCMC and Bootstrap uncertainty estimation on each of our tasks. We see that both techniques yield fairly similar confidence intervals across the various tasks.

In the case of APPS, we observe a marginally wider confidence interval using bootstrapping. If we convert these loss-based 90\% confidence intervals on APPS into parameter-count (under the LLaMA 2 scaling law) based confidence intervals, we see that the wider bootstrapping interval corresponds to a range of 225B to 830B parameters rather than 244B to 509B parameters with MCMC.

Our MCMC posterior sampling uses a uniform prior over parameters. We sample using the No-U-Turn Sampler~\cite{hoffman2014no} implemented in Numpyro~\citep{phan2019composable}. We use 4 chains with 25k samples each and 15k warmup steps each, totaling 100k samples. We initialize each chain using the maximum likelihood parameters found using the procedure described in Appendix~\ref{app:fitting}. We find that without tuning the temperature of the energy function (e.g., $\text{Energy} = \text{loss}/\text{temperature}$) the sampler is extremely unstable, demonstrating wildly out-of-distribution posterior samples. To control this instability, we sweep over temperature values 1e-3, 1e-4, 1e-5, 1e-6, 1e-7, 1e-8, 1e-9 and select the greatest such temperature for which the mode of the sampled distribution is centered around the maximum likelihood estimate.

To obtain the bootstrap uncertainty estimate, we fit our emergence law on 1000 bootstrap samples of the data. Since our emergence law fitting procedure weighs each example in inverse proportion to the size of the finetuning data amount $D$ (see Section~\ref{sec:collecting_empirical_data}), we perform weighted bootstrap sampling by sampling with replacement from a distribution proportional to each datapoint's weight. The number of datapoints we sample for each bootstrap set is equal to the total sum of the dataset weights. We then count the number of times each datapoint is sampled and use this count as the new weight when fitting the emergence law on the bootstrap dataset.

\begin{table}[h]
    \centering
    \scriptsize
    \setlength{\tabcolsep}{4pt}
    \setlength{\lightrulewidth}{0.01em}  
    \setlength{\heavyrulewidth}{0.175em}  
    \begin{tabular}{ll|rrrrrrrr|r}
        \toprule
        \textbf{Task} & \textbf{Method} & \textbf{5\%} & \textbf{10\%} & \textbf{25\%} & \textbf{50\%} & \textbf{75\%} & \textbf{90\%} & \textbf{95\%} & \textbf{MLE} & \textbf{GT} \\
        \midrule[\heavyrulewidth]
        \addlinespace[0.5em]
        \multirow{2}{*}{GSM8K} & MCMC & 1.813 & 1.852 & 1.900 & 1.937 & 1.970 & 1.992 & 2.003 & \multirow{2}{*}{2.006} & \multirow{2}{*}{1.984} \\
        & Bootstrap & 1.978 & 1.984 & 1.995 & 2.007 & 2.021 & 2.031 & 2.036 & & \\
        \midrule[\lightrulewidth]
        \multirow{2}{*}{MMLU} & MCMC & 1.825 & 1.828 & 1.837 & 1.847 & 1.858 & 1.866 & 1.869 & \multirow{2}{*}{1.855} & \multirow{2}{*}{1.814} \\
        & Bootstrap & 1.818 & 1.825 & 1.836 & 1.848 & 1.859 & 1.867 & 1.871 & & \\
        \midrule[\lightrulewidth]
        \multirow{2}{*}{CSQA} & MCMC & 1.781 & 1.810 & 1.821 & 1.829 & 1.835 & 1.840 & 1.843 & \multirow{2}{*}{1.830} & \multirow{2}{*}{1.827} \\
        & Bootstrap & 1.723 & 1.736 & 1.815 & 1.835 & 1.846 & 1.857 & 1.863 & & \\
        \midrule[\lightrulewidth]
        \multirow{2}{*}{CoLA} & MCMC & 1.712 & 1.724 & 1.742 & 1.761 & 1.779 & 1.795 & 1.804 & \multirow{2}{*}{1.769} & \multirow{2}{*}{1.833} \\
        & Bootstrap & 1.738 & 1.746 & 1.758 & 1.770 & 1.782 & 1.791 & 1.798 & & \\
        \midrule[\heavyrulewidth]
        \addlinespace[0.5em]
        \multirow{2}{*}{MMLU C4 V1} & MCMC & 2.207 & 2.221 & 2.241 & 2.246 & 2.255 & 2.259 & 2.261 & \multirow{2}{*}{2.254} & \multirow{2}{*}{2.226} \\
        & Bootstrap & 2.183 & 2.200 & 2.216 & 2.228 & 2.238 & 2.246 & 2.250 & & \\
        \midrule[\lightrulewidth]
        \multirow{2}{*}{MMLU C4 V2} & MCMC & 2.264 & 2.275 & 2.289 & 2.306 & 2.310 & 2.316 & 2.320 & \multirow{2}{*}{2.311} & \multirow{2}{*}{2.318} \\
        & Bootstrap & 2.249 & 2.257 & 2.272 & 2.284 & 2.296 & 2.305 & 2.311 & & \\
        \midrule[\heavyrulewidth]
        \addlinespace[0.5em]
        \multirow{2}{*}{APPS} & MCMC & 1.324 & 1.332 & 1.344 & 1.357 & 1.370 & 1.380 & 1.386 & \multirow{2}{*}{1.361} & \multirow{2}{2em}{\centering ---} \\
        & Bootstrap & 1.285 & 1.304 & 1.330 & 1.352 & 1.370 & 1.385 & 1.393 & & \\
        \bottomrule
    \end{tabular}
    \caption{\footnotesize{Comparing emergence prediction uncertainty estimates obtained via MCMC and bootstrapping. On each task, we present seven a range of percentiles for the point of emergence in terms of pretraining loss for each distribution. We also present the maximum likelihood prediction (``MLE''), and the ground-truth (``GT'') point of emergence. We see that both methods generally produce similar distributions. In the top section we present the uncertainties for each task used in Section~\ref{sec:full_fits_main_paper}. In the middle we include uncertainties for the data quality experiments in Section~\ref{sec:data_quality}. ``MMLU C4 V1'' refers to the OpenLLaMA V1 fit and ``MMLU C4 V2'' refers to the V2 fit. At the bottom, we include uncertainties for the APPS experiment in Section~\ref{sec:future_models}.}}
    \label{tab:uncertainty_ablation_results}
\end{table}

\subsection{\textbf{OpenLLaMA V1 Verses V2 Pretraining Data}}
\label{app:emergence_openllama_v2}

The OpenLLaMA V1 models were trained on the RedPajama~\citep{together2023redpajama} dataset, whereas the V2 models were trained on a custom mixture of the Falcon Refined-web~\citep{refinedweb} dataset, the StarCoder~\citep{li2023starcoder} dataset, and the RedPajama~\citep{together2023redpajama} dataset. The V2 models also used a different tokenizer, which was trained on its corresponding pretraining dataset. Most hyperparameters between the two series of models remain the same.

\subsection{\textbf{LLaMA 2 Scaling Law Details}}
\label{app:llama2_scaling_law_details}

\begin{figure*}
    \includegraphics[width=0.99\textwidth]{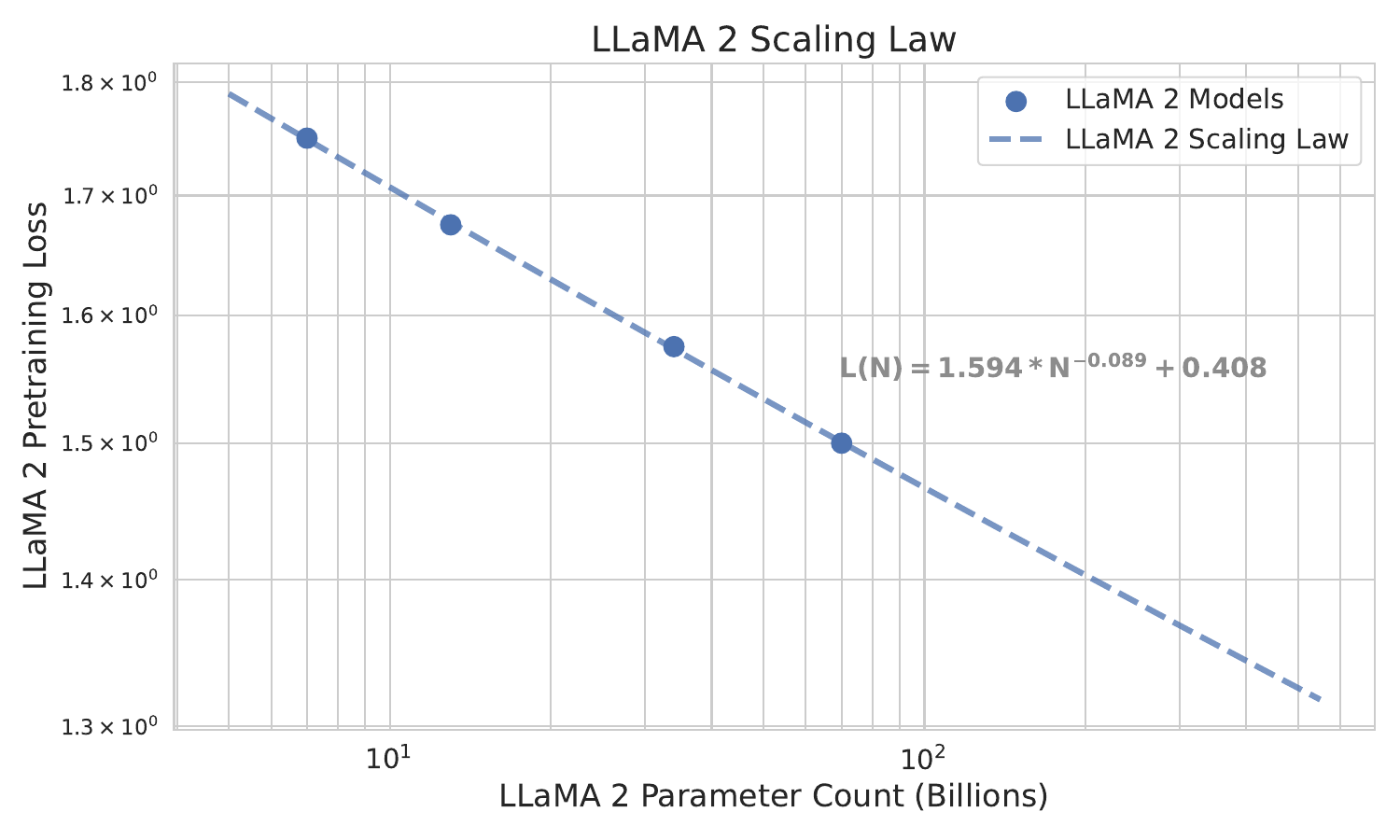}
    \caption{\footnotesize{We plot our scaling law fit for the LLaMA 2 series of models. We also include the learned values for our final fit on the plot. In this case $N$ corresponds to parameter count in billions. We see that the LLaMA 2 models are well modeled by our scaling law.}}
    \label{fig:llama2_scaling_law}
\end{figure*}

We provide details for our LLaMA 2 scaling law fit in Section~\ref{sec:future_models}. We use the LLaMA 2 7B, 13B, 34B, and 70B models to fit the scaling law. We extract the loss for each model from Figure 5 in~\cite{touvron2023llama2openfoundation}. Specifically we use losses of 1.75, 1.675, 1.575, and 1.5 nats for the 7B, 13B, 34B, and 70B models respectively. All models were trained on the same 2 trillion tokens. Therefore, since the data amount is fixed, we modify the functional form from~\cite{hoffmann2022training} by absorbing the data scaling term into the irreducible loss (e.g., $E$). This gives a functional form of: a $L(N) = \frac{A}{N^\alpha}+E$. In this case $A$, $\alpha$, and $E$ are all learned parameters, and $N$ corresponds to parameter count in billions. We include a plot of our LLaMA 2 scaling law fit in Figure~\ref{fig:llama2_scaling_law}.

To fit the scaling law we follow a similar procedure to~\cite{hoffmann2022training}. We optimize mean squared error in log space. We first perform a brute-force grid-search over all the values in: $A \in \{0.0, 0.1, 0.2, ..., 5.0\}$, $e^{\alpha} \in \{0.0, 0.01, 0.02, ..., 1.0\}$, $E \in \{0.0, 0.01, 0.02, ..., 1.0\}$. We select the 100k parameters from this grid search with the lowest loss and then run the L-BFGS optimizer initialized from each one. We then select the best final fit from all optimization runs.

\subsection{\textbf{Additional Experiment Details}}
\label{app:details}

For all experiments, to select the set of finetuning data subsets used for fitting our emergence law, we first conduct a logarithmic sweep over various data amounts (e.g., $\frac{1}{2}$ the data, $\frac{1}{4}$ the data, $\frac{1}{8}$ the data, etc...) and then collect additional subsets nearer to the limit at which emergence is visible with our 3B checkpoints. To account for noise in the data sub-sampling process, we finetune on two different sampled subsets for each data subset level.

For each experiment with OpenLLaMA V1 and V2, we use at least 6 model checkpoints for fitting the emergence law. In all tasks we used the same initial set of 6 checkpoints. However, on some tasks, we found that after finetuning on the full data, the earliest of the 6 checkpoints was significantly post-emergence. We therefore used up to 3 additional earlier checkpoints in these cases. We include full details of which checkpoints we used for each task below. All checkpoints are publically available at \url{https://huggingface.co/openlm-research}.

For, all tasks we perform full parameter finetuning using the AdamW optimizer. We use 0.05 dropout, 0.01 weight decay, Adam $\beta_1=0.9$, $\beta_2=0.95$, and learning rate schedule with linear warmup followed by a constant learning rate. We hold out 10\% of our training data for validation. We perform a small sweep over learning rate, batch size, and the number of learning-rate warm-up steps, selecting the best setting that produces the best validation loss. We perform early stopping according to validation loss on all tasks except on CoLA, which we describe in more detail below. For all of our experiments, except those in Section~\ref{sec:data_quality} and Section~\ref{sec:future_models}, we use the OpenLLaMA V1~\citep{openlm2023openllama} models that were pretrained on 1T tokens from the Red-pajama dataset~\citep{together2023redpajama}. In Section~\ref{sec:data_quality}, we also use the OpenLLaMA V2 models which were pretrained on a different corpus, also for $\sim$1T tokens. Finally, in Section~\ref{sec:future_models}, we use the LLaMA 2 series of models~\citep{touvron2023llama2openfoundation} (7B, 13B, and 70B), which were pretrained for 2T tokens on a proprietary corpus. Below we include more details on our experiments for each task.

\paragraph{MMLU.} For predicting emergence on MMLU we use 6 intermediate 3B model checkpoints pretrained for 42B, 210B, 419B, 629B, 839B, and 1T tokens. We train on the MMLU test set and then evaluate on the validation set. In addition to the full data, we train each checkpoint on two different randomly sampled dataset subsets of each fraction: $\frac{1}{2}$, $\frac{1}{4}$, $\frac{1}{8}$, $\frac{1}{16}$, $\frac{3}{16}$, and $\frac{3}{32}$. We use learning rate 5e-6, batch size 256, and 24 learning-rate warmup steps. We use the standard 5-shot LM evaluation harness~\citep{touvron2023llama} prompt for our few-shot results. For the OpenLLaMA V2 checkpoints in Section~\ref{sec:data_quality}, we use model checkpoints pretrained for 42B, 210B, 419B, 629B, 839B, and 965B tokens. Otherwise, use these same settings for our experiment with OpenLLaMA V2.

\paragraph{GSM8K.} For predicting emergence on GSM8K we use 8 intermediate 3B model checkpoints pretrained for 21B, 31B, 42B, 210B, 419B, 629B, 839B, and 1T tokens. We train and test on the standard splits. On this task our models are trained to output a chain-of-thought~\citep{wei2022chain} followed by a final answer, using the CoT traces provided with the GSM8K dataset. In addition to the full data, we train each checkpoint on two different randomly sampled dataset subsets of each fraction: $\frac{1}{2}$, $\frac{1}{4}$, $\frac{1}{8}$, $\frac{1}{16}$, $\frac{1}{32}$, and $\frac{3}{64}$. We use learning rate of 1e-5, a batch size of 32, and no learning-rate warmup steps. For our prompted evaluation, we use the 6-shot chain-of-thought prompt from~\cite{zelikman2022star}.

\paragraph{CommonsenseQA.} For predicting emergence on CommonsenseQA we use 9 intermediate 3B model checkpoints pretrained for 10B, 21B, 31B, 42B, 210B, 419B, 629B, 839B, and 1T tokens. We train on the standard train set and evaluate on the validation set. In addition to the full data, we train each checkpoint on two different randomly sampled dataset subsets of each fraction: $\frac{1}{2}$, $\frac{1}{4}$, $\frac{1}{8}$, $\frac{1}{16}$, $\frac{1}{32}$, $\frac{3}{64}$, $\frac{3}{32}$, and $\frac{3}{16}$. We use a learning rate of 5e-6, a batch size of 64, and 96 learning-rate warmup steps. For our prompted evaluation we use the 7-shot prompt from~\cite{wei2022chain} with the chain of thought examples stripped from the prompt, just showing the final answer.

\paragraph{CoLA.} For predicting emergence on CoLA we use 9 intermediate 3B model checkpoints pretrained for 10B, 21B, 31B, 42B, 210B, 419B, 629B, 839B, and 1T tokens. We train on the standard train set and evaluate on the validation set. In addition to the full data, we train each checkpoint on two different randomly sampled dataset subsets of each fraction: $\frac{1}{2}$, $\frac{1}{4}$, $\frac{1}{8}$, $\frac{1}{16}$, $\frac{1}{32}$, $\frac{1}{64}$, $\frac{1}{128}$, $\frac{3}{64}$, $\frac{3}{128}$, and $\frac{3}{256}$. We use a learning rate of 5e-6, a batch size of 256, and 48 learning-rate warmup steps. For our prompted evaluation we use the standard 5-shot prompt in the LM evaluation harness~\citep{touvron2023llama}. On CoLA we deviate slightly from our early stopping procedure. We find that for some of the smaller data subsets, downstream validation performance continues to improve after overfitting. In these cases we perform early stopping according to the downstream MCC evaluation on the validation set rather than the validation loss.

\paragraph{APPS.} For predicting emergence on APPS we use the 7B, 13B, and 70B LLaMA 2 base models~\citep{touvron2023llama2openfoundation}, each of which were trained for 2T tokens. In addition to the full data, we train each model on two different randomly sampled dataset subsets of each fraction: $\frac{1}{2}$, $\frac{1}{4}$, $\frac{1}{8}$, $\frac{1}{16}$, $\frac{1}{32}$, $\frac{1}{64}$, $\frac{1}{128}$, $\frac{1}{256}$, and $\frac{3}{128}$. In the APPS training set there are multiple solutions provided for each answer, we include all solutions in our finetuning dataset. However, when we sample subsets (e.g., $\frac{1}{2}$ the data), rather than taking a random sample of all solutions, we take a random sample of the questions, and include all solutions per question. We use a learning rate of 5e-6, a batch size of 64, and no learning-rate warmup steps. For our prompted evaluation we use the 2-shot prompt in Appendix~\ref{app:apps_prompt}.

All finetuning experiments were conducted on TPU-V3 and TPU-V5e pods using JAX~\citep{jax2018github} and Scalax~\citep{scalax2024github} for distributed training.

\subsection{\textbf{Full Emergence Prediction Plots}}
\label{app:full_plots}

\begin{figure*}
    \centering
    \includegraphics[width=0.99\textwidth]{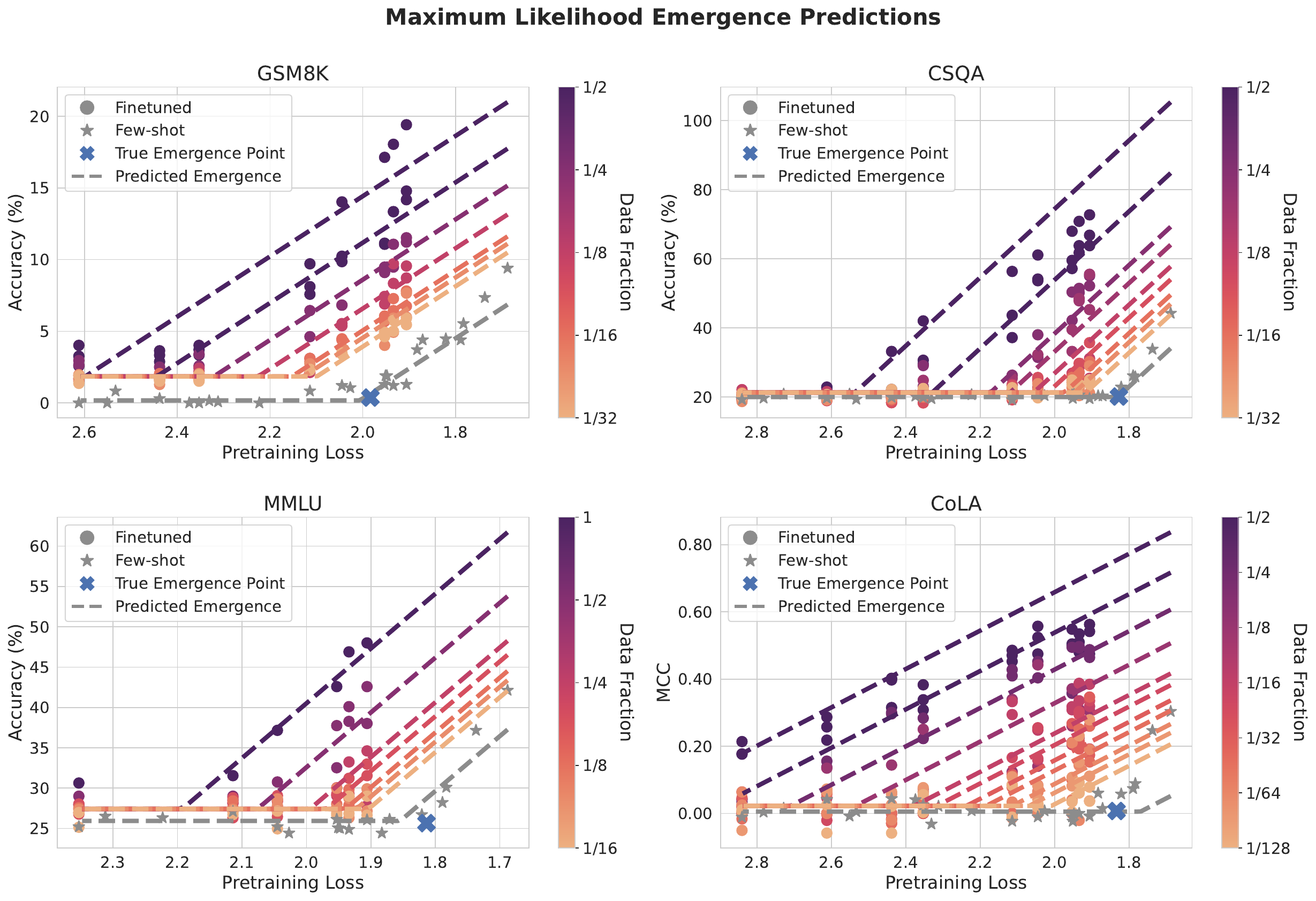}
    \caption{\footnotesize{We plot the maximum likelihood predictions from our emergence law on each task. These plots include results from every finetuning run used for fitting the emergence law. The grey line represents our extrapolated prediction and the multi-color lines correspond to the fit produced by the emergence law for the various data levels. We see that across all tasks we are able to successfully predict the point of emergence within 0.1 nats and in many cases much less than that.}}
    \label{fig:full_fits}
\end{figure*}

In Figure~\ref{fig:full_fits} we plot all of the data used for our maximum likelihood emergence predictions from Section~\ref{sec:evaluating_emergence}.

\begin{figure*}
    \centering
    \includegraphics[width=0.99\textwidth]{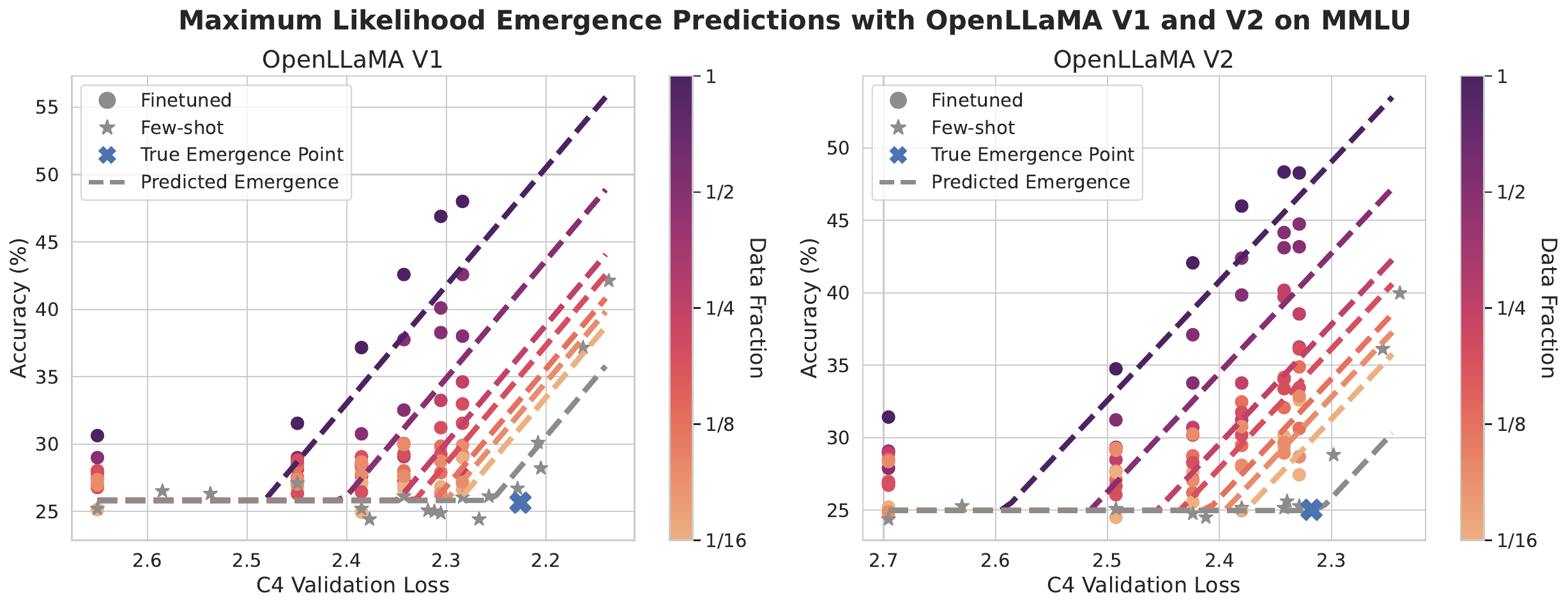}
    \caption{\footnotesize{We plot the maximum likelihood predictions from our emergence law with OpenLLaMA V1 (left) and OpenLLaMA V2 (right) on MMLU. We plot C4 Validation loss on the x-axis. These plots include results from every finetuning run used for fitting the emergence law. The grey line represents our extrapolated prediction and the multi-color lines correspond to the fit produced by the emergence law for the various data levels. We see that in both cases we are able to successfully predict the point of emergence within 0.1 nats.}}
    \label{fig:full_fits_mmlu_v1_v2}
\end{figure*}

In Figure~\ref{fig:full_fits_mmlu_v1_v2} we plot all the data for our maximum likelihood emergence predictions using OpenLLaMA V1 and V2 on MMLU in Section~\ref{sec:data_quality}.

\begin{figure*}
    \centering
    \subfigure{
        \includegraphics[width=0.99\textwidth]{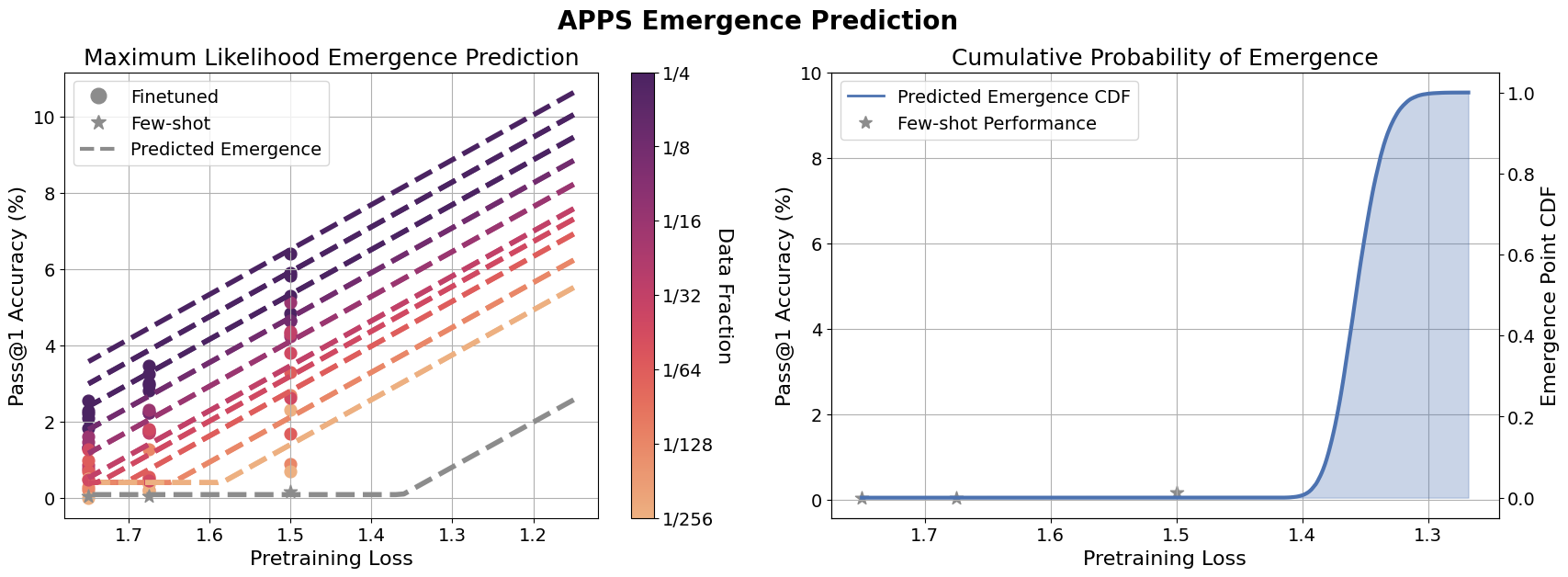}
        \label{fig:apps_all}
    }
    \caption{\footnotesize{We plot the MLE prediction (left) and MCMC CDF (right) for our emergence law fit using LLaMA 2 on APPS. The left plot includes results from every finetuning run used for fitting the emergence law. The grey line represents our extrapolated prediction and the multi-color lines correspond to the fit produced by the emergence law for the various data levels. We see that our emergence law predicts emergence roughly 0.15 nats beyond the LLaMA 2 70B model.}}
    \label{fig:full_fits_apps}
\end{figure*}

In Figure~\ref{fig:full_fits_apps} we plot all the data for our MLE prediction and MCMC CDF using LLaMA 2 on APPS in Section~\ref{sec:future_models}.

\begin{figure*}
    \centering
    \includegraphics[width=0.99\textwidth]{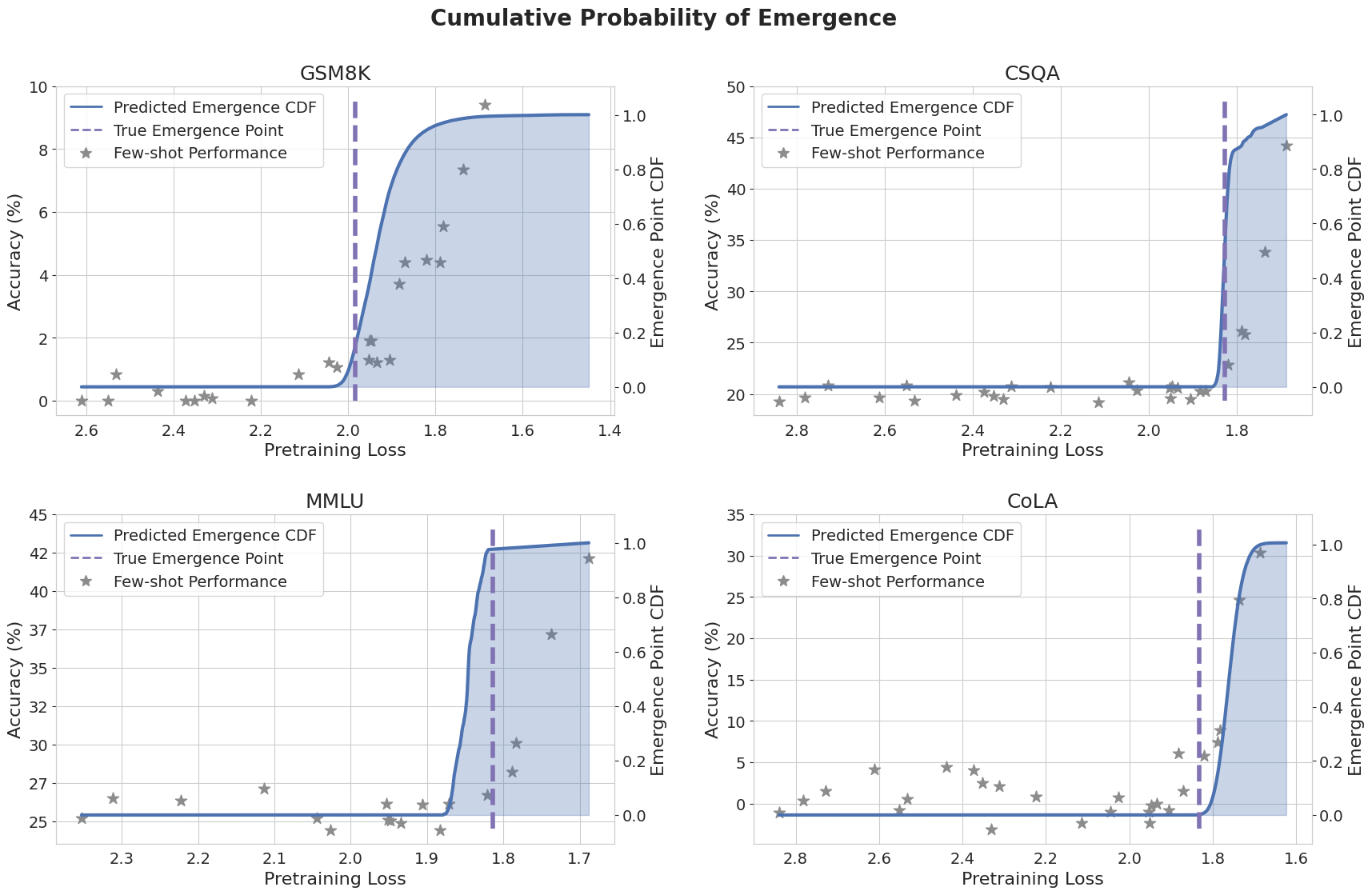}
    \caption{\footnotesize{We plot the cumulative distribution function of our estimated posterior distribution over the point of emergence on each task. The stars correspond to few-shot performance on the task and represent the true emergence curve. The point at which the slope of the CDF peaks represents the mode of the distribution. We see across all tasks that the distribution spikes near the true point of emergence and is followed by a moderately long tail.}}
    \label{fig:all_mcmc}
\end{figure*}

Finally, in Figure~\ref{fig:all_mcmc}, we plot the MCMC CDF of the point of emergence under our emergence law for all tasks in Section~\ref{sec:evaluating_emergence}.

\newpage

\subsection{\textbf{APPS 2-shot Prompt}}
\label{app:apps_prompt}

We include the two shot prompt we use for few-shot evaluation of LLaMA 2 on APPS below:

\begin{lstlisting}
"""
You are given an array $a$ of length $n$ consisting of zeros. You perform $n$ actions with this array: during the $i$-th action, the following sequence of operations appears:  Choose the maximum by length subarray (continuous subsegment) consisting only of zeros, among all such segments choose the leftmost one;  Let this segment be $[l; r]$. If $r-l+1$ is odd (not divisible by $2$) then assign (set) $a[\frac{l+r}{2}] := i$ (where $i$ is the number of the current action), otherwise (if $r-l+1$ is even) assign (set) $a[\frac{l+r-1}{2}] := i$. 

Consider the array $a$ of length $5$ (initially $a=[0, 0, 0, 0, 0]$). Then it changes as follows:  Firstly, we choose the segment $[1; 5]$ and assign $a[3] := 1$, so $a$ becomes $[0, 0, 1, 0, 0]$;  then we choose the segment $[1; 2]$ and assign $a[1] := 2$, so $a$ becomes $[2, 0, 1, 0, 0]$;  then we choose the segment $[4; 5]$ and assign $a[4] := 3$, so $a$ becomes $[2, 0, 1, 3, 0]$;  then we choose the segment $[2; 2]$ and assign $a[2] := 4$, so $a$ becomes $[2, 4, 1, 3, 0]$;  and at last we choose the segment $[5; 5]$ and assign $a[5] := 5$, so $a$ becomes $[2, 4, 1, 3, 5]$. 

Your task is to find the array $a$ of length $n$ after performing all $n$ actions. Note that the answer exists and unique.

You have to answer $t$ independent test cases.


-----Input-----

The first line of the input contains one integer $t$ ($1 \le t \le 10^4$) - the number of test cases. Then $t$ test cases follow.

The only line of the test case contains one integer $n$ ($1 \le n \le 2 \cdot 10^5$) - the length of $a$.

It is guaranteed that the sum of $n$ over all test cases does not exceed $2 \cdot 10^5$ ($\sum n \le 2 \cdot 10^5$).


-----Output-----

For each test case, print the answer - the array $a$ of length $n$ after performing $n$ actions described in the problem statement. Note that the answer exists and unique.


-----Example-----
Input
6
1
2
3
4
5
6

Output
1 
1 2 
2 1 3 
3 1 2 4 
2 4 1 3 5 
3 4 1 5 2 6
"""
from collections import defaultdict as dd
from collections import deque
import bisect
import heapq

def ri():
    return int(input())

def rl():
    return list(map(int, input().split()))


def solve():
    n = ri()
    output = [0] * (n)

    Q = [(-n, 0 ,n - 1)]
    for i in range(1, n + 1):
        prev = heapq.heappop(Q)
        lo, hi = prev[1], prev[2]
        mid = (lo + hi) // 2
        output[mid] = i
        if mid > lo:
            heapq.heappush(Q, (-(mid - 1 - lo), lo, mid - 1))
        if hi > mid:
            heapq.heappush(Q, (-(hi - 1 - mid), mid + 1, hi))
    print(*output)





mode = 'T'

if mode == 'T':
    t = ri()
    for i in range(t):
        solve()
else:
    solve()

"""
$n$ robots have escaped from your laboratory! You have to find them as soon as possible, because these robots are experimental, and their behavior is not tested yet, so they may be really dangerous!

Fortunately, even though your robots have escaped, you still have some control over them. First of all, you know the location of each robot: the world you live in can be modeled as an infinite coordinate plane, and the $i$-th robot is currently located at the point having coordinates ($x_i$, $y_i$). Furthermore, you may send exactly one command to all of the robots. The command should contain two integer numbers $X$ and $Y$, and when each robot receives this command, it starts moving towards the point having coordinates ($X$, $Y$). The robot stops its movement in two cases:  either it reaches ($X$, $Y$);  or it cannot get any closer to ($X$, $Y$). 

Normally, all robots should be able to get from any point of the coordinate plane to any other point. Each robot usually can perform four actions to move. Let's denote the current coordinates of the robot as ($x_c$, $y_c$). Then the movement system allows it to move to any of the four adjacent points:  the first action allows it to move from ($x_c$, $y_c$) to ($x_c - 1$, $y_c$);  the second action allows it to move from ($x_c$, $y_c$) to ($x_c$, $y_c + 1$);  the third action allows it to move from ($x_c$, $y_c$) to ($x_c + 1$, $y_c$);  the fourth action allows it to move from ($x_c$, $y_c$) to ($x_c$, $y_c - 1$). 

Unfortunately, it seems that some movement systems of some robots are malfunctioning. For each robot you know which actions it can perform, and which it cannot perform.

You want to send a command so all robots gather at the same point. To do so, you have to choose a pair of integer numbers $X$ and $Y$ so that each robot can reach the point ($X$, $Y$). Is it possible to find such a point?


-----Input-----

The first line contains one integer $q$ ($1 \le q \le 10^5$) - the number of queries.

Then $q$ queries follow. Each query begins with one line containing one integer $n$ ($1 \le n \le 10^5$) - the number of robots in the query. Then $n$ lines follow, the $i$-th of these lines describes the $i$-th robot in the current query: it contains six integer numbers $x_i$, $y_i$, $f_{i, 1}$, $f_{i, 2}$, $f_{i, 3}$ and $f_{i, 4}$ ($-10^5 \le x_i, y_i \le 10^5$, $0 \le f_{i, j} \le 1$). The first two numbers describe the initial location of the $i$-th robot, and the following four numbers describe which actions the $i$-th robot can use to move ($f_{i, j} = 1$ if the $i$-th robot can use the $j$-th action, and $f_{i, j} = 0$ if it cannot use the $j$-th action).

It is guaranteed that the total number of robots over all queries does not exceed $10^5$.


-----Output-----

You should answer each query independently, in the order these queries appear in the input.

To answer a query, you should do one of the following:  if it is impossible to find a point that is reachable by all $n$ robots, print one number $0$ on a separate line;  if it is possible to find a point that is reachable by all $n$ robots, print three space-separated integers on the same line: $1$ $X$ $Y$, where $X$ and $Y$ are the coordinates of the point reachable by all $n$ robots. Both $X$ and $Y$ should not exceed $10^5$ by absolute value; it is guaranteed that if there exists at least one point reachable by all robots, then at least one of such points has both coordinates not exceeding $10^5$ by absolute value.


-----Example-----
Input
4
2
-1 -2 0 0 0 0
-1 -2 0 0 0 0
3
1 5 1 1 1 1
2 5 0 1 0 1
3 5 1 0 0 0
2
1337 1337 0 1 1 1
1336 1337 1 1 0 1
1
3 5 1 1 1 1

Output
1 -1 -2
1 2 5
0
1 -100000 -100000
"""
def main():
    import sys
    input = sys.stdin.readline
    
    def solve():
        n = int(input())
        maxx = 10**5
        minx = -10**5
        maxy = 10**5
        miny = -10**5
        
        for _ in range(n):
            x, y, f1, f2, f3, f4 = map(int, input().split())
            if not f1:
                minx = max(minx, x)
            if not f2:
                maxy = min(maxy, y)
            if not f3:
                maxx = min(maxx, x)
            if not f4:
                miny = max(miny, y)
        
        if minx > maxx or miny > maxy:
            print(0)
        else:
            print(1, minx, miny)
        
    
    for _ in range(int(input())):
        solve()
    
    return 0

main()

"""
{QUESTION}
"""
\end{lstlisting}

\end{document}